\documentclass[preprint, review, 3p, authoryear, times]{elsarticle}

\usepackage{enumitem}
\usepackage{placeins} 

\usepackage{caption}
\usepackage{float} 
\usepackage{graphicx}
\usepackage{url}
\usepackage{color, soul}
\usepackage{bm}
\usepackage{amsmath} 
\usepackage{lineno}

\begin{document}


\title{
From PowerPoint UI Sketches to Web-Based Applications: Pattern-Driven Code Generation for GIS Dashboard Development Using Knowledge-Augmented LLMs, Context-Aware Visual Prompting, and the React Framework
} 


\tnotetext[t1]{This manuscript has been co-authored by UT-Battelle, LLC, under contract DE-AC05-00OR22725 with the US Department of Energy (DOE). The US government retains and the publisher, by accepting the article for publication, acknowledges that the US government retains a nonexclusive, paid-up, irrevocable, worldwide license to publish or reproduce the published form of this manuscript, or allow others to do so, for US government purposes. DOE will provide public access to these results of federally sponsored research in accordance with the DOE Public Access Plan (http://energy.gov/downloads/doe-public-access-plan).}
\author[CSED]{Haowen Xu}\ead{xuh4@ornl.gov}
\author[MSTD]{Xiao-Ying Yu \corref{cor1}} \ead{}

\cortext[cor1]{Corresponding author.}

\address[CSED]{Computational Urban Sciences Group, Oak Ridge National Laboratory, Oak Ridge, TN 37830, USA}
\address[MSTD]{Materials Science and Technology Division, Oak Ridge National Laboratory, Oak Ridge, TN 37830, USA}


\makeatletter
\newcommand{\printfnsymbol}[1]{%
  \textsuperscript{\@fnsymbol{#1}}%
}

\newcommand*{\MyIndent}{\hspace*{0.5cm}}%


\begin{abstract} \label{sec:abstract}
Developing web-based GIS applications, commonly known as CyberGIS dashboards, for querying and visualizing GIS data in environmental research often demands repetitive and resource-intensive efforts. While Generative AI offers automation potential for code generation, it struggles with complex scientific applications due to challenges in integrating domain knowledge, software engineering principles, and UI design best practices.
This paper introduces a knowledge-augmented code generation framework that retrieves software engineering best practices, domain expertise, and advanced technology stacks from a specialized knowledge base to enhance Generative Pre-trained Transformers (GPT) for front-end development. The framework automates the creation of GIS-based web applications (e.g., dashboards, interfaces) from user-defined UI wireframes sketched in tools like PowerPoint or Adobe Illustrator. A novel Context-Aware Visual Prompting method, implemented in Python, extracts layouts and interface features from these wireframes to guide code generation.
Our approach leverages Large Language Models (LLMs) to generate front-end code by integrating structured reasoning, software engineering principles, and domain knowledge, drawing inspiration from Chain-of-Thought (CoT) prompting and Retrieval-Augmented Generation (RAG). A case study demonstrates the framework’s capability to generate a modular, maintainable web platform hosting multiple dashboards for visualizing environmental and energy data (e.g., time-series, shapefiles, rasters) from user-sketched wireframes.
By employing a knowledge-driven approach, the framework produces scalable, industry-standard front-end code using design patterns such as Model-View-ViewModel (MVVM) and frameworks like React. This significantly reduces manual effort in design and coding, pioneering an automated and efficient method for developing smart city software.


\end{abstract} \label{sec:abstract}
 \maketitle


\section{Introduction}
\label{Introduction}
Over the past decades, scientific web applications (Web Apps), such as CyberGIS systems, visual analytics dashboards, digital twin platforms, and online decision support systems, have become indispensable tools for both the scientific community and the public to discover, query, visualize, and download vast urban and environmental data sets for smart city research \citep{ferre2022adoption, dembski2020urban}. Aligned with the NSF's cyberinfrastructure (CI) initiative, academia and government agencies have increasingly adopted these applications as part of interdisciplinary informatics projects, providing effective, user-friendly tools for data dissemination \citep{yu2021coevolution}. With advances in internet and communication technologies, computing hardware, and artificial intelligence, these tools are transforming urban and environmental research by enabling data- and simulation-driven insights for decision support \citep{kadupitige2022enhancing}, fostering collaborative research through data and simulation integration \citep{parashar2019virtual} and enhancing education and public engagement in citizen science and voluntary data collection \citep{skarlatidou2019volunteers}. Key application areas include water resource management \citep{souffront2018cyberinfrastructure, xu2022overview}, hazard mitigation \citep{mandal2024prime, xu2020web, garg2018cloud}, intelligent transportation systems \citep{xu2023smart, xu2022interactive, ghosh2017intelligent}, connected and automated vehicles \citep{xu2023mobile, kampmann2019dynamic}, built-environment and building energy management \citep{jia2019adopting, kim2022design, xu2022geo}, pandemic management \citep{xu2021episemblevis, li2021emerging, thakur2020covid}, and urban planning and design \citep{alatalo2017two}. Numerous interdisciplinary studies highlight the transformative potential of these web applications in advancing environmental and urban research, as well as smart city management. Their ongoing evolution is driven by the integration of emerging technologies like artificial intelligence, the Internet of Things (IoT), edge computing, and cyber-physical systems.

Despite advancements in scientific web applications, developing customized tools like cyberGIS and digital twin platforms for integrating and visualizing diverse environmental or urban data (e.g., hydrological, traffic flow, meteorological data, or simulations) remains highly demanding and resource-intensive \citep{shanjun2024design, siddiqui2024digital, lei2023challenges}. These efforts require expertise in software and data engineering, as well as time-consuming tasks like client-server development, database management, real-time analytics, machine learning, and simulations \citep{ikegwu2022big}. Consequently, researchers often shift from their core scientific work to learn complex web programming, UI/UX design, and database technologies \citep{li2022bibliometric}. Although modern software practices like design patterns aim to streamline development, their effective use demands specialized software engineering knowledge \citep{fayad2015software}. Researchers with data analytics expertise often lack formal software development experience, facing challenges even when skilled engineers are involved \citep{kim2017data}. Designing, deploying, and maintaining large-scale web apps remains labor-intensive, limiting scalability and adaptability in platforms like cyberGIS or digital twins \citep{shah2024optimizing, mcbreen2002software, liu2015cybergis}. Emerging Generative AI (GenAI) technologies offer potential to automate web development for environmental and urban research. While studies have shown the feasibility of training large language models (LLMs) like GPT for automating data analytics and development tasks \citep{liang2024can, liukko2024chatgpt}, challenges remain for complex scientific web apps due to inefficiencies in prompting methods, limitations in domain knowledge, and LLMs' attention mechanisms, which are trained on generalized text data.

\begin{figure*}[htb]
 \centering
\includegraphics[width=\textwidth]{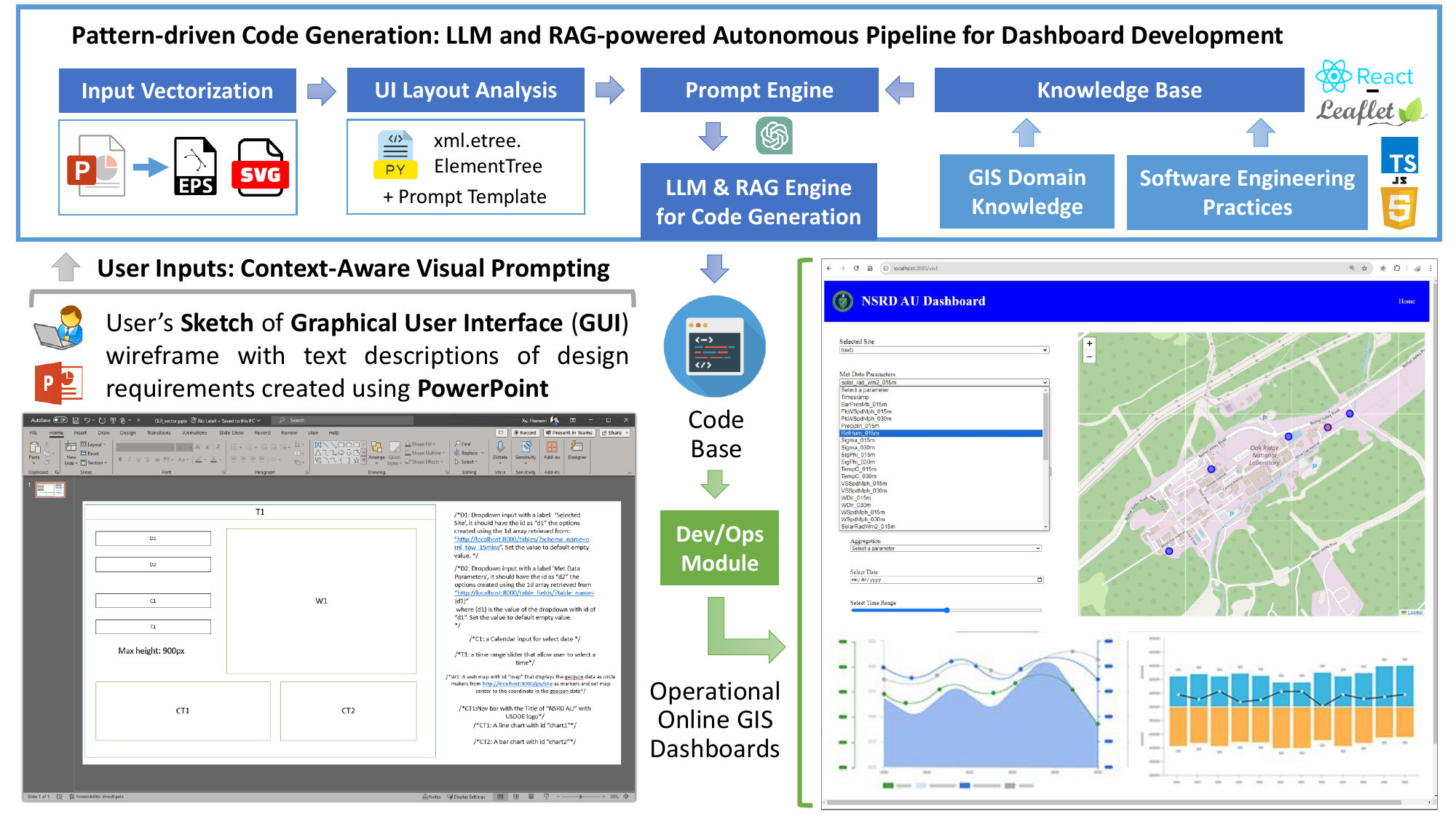}
 \caption{From annotated wireframe to code generation, a knowledge driven framework for automated software development of cyberGIS platform for visualizing time-series and spatial data. }
 \label{fig:concepts}
\end{figure*}

This paper introduces a knowledge-augmented code generation framework that integrates software engineering best practices, domain expertise, and advanced software stacks to enhance Generative Pre-trained Transformers (GPT) for front-end development. The framework automates the generation of GIS-based web applications, including visualization dashboards and analytical interfaces, directly from user-defined UI wireframes created in tools like PowerPoint or Adobe Illustrator. It leverages industry-standard web frameworks and software engineering best practices to ensure scalability, maintainability, and efficiency. We propose a novel context-Aware visual prompting method, implemented in Python, that interprets wireframes to extract layout structures and interface elements, enabling LLMs to generate front-end code while incorporating both software engineering principles and domain knowledge through Chain-of-Thought (CoT) and Retrieval-Augmented Generation (RAG) paradigms. A case study demonstrates the framework’s capability to autonomously generate a modular and maintainable web platform hosting multiple dashboards for visualizing and interacting with environmental and energy data, including time-series data sets, GIS shapefiles, and raster imagery, based on user-sketched wireframes. By leveraging a knowledge-driven approach, the framework ensures that the generated front-end code adheres to industry-standard software design and architectural patterns, including the React framework and Model-View-ViewModel (MVVM) architecture, facilitating long-term scalability and maintainability. This approach significantly reduces manual effort in UI/UX design, coding, and maintenance, pioneering a scalable, autonomous, and robust solution for developing web applications that support smart city advancements.

\section{Literature Review}
\label{Literature Review}
LLMs like GPT-3 \citep{brown2020language}, GPT-4 \citep{sun2023gpt}, and DeepSeek \citep{guo2024deepseek} have revolutionized software development through autonomous code generation. Their ability to understand and generate human language enables tasks like reasoning, code generation, and problem-solving \citep{li2023autonomous, tupayachi2024towards}. Integrating LLMs with RAG techniques enhances domain-specific knowledge retrieval, improving code accuracy and reliability. This synergy automates key development processes, including requirement definition, bug fixing, and program repair, reducing manual effort. Ultimately, LLMs and RAG offer scalable solutions for automating repetitive and complex coding tasks \citep{meyer2023llm, baldazzi2023fine}.

\subsection{A Review of LLM-Based Approaches in Software Development}
LLMs are increasingly applied in diverse software engineering tasks, particularly in code generation. This paper focuses on evaluating their capabilities in web application development. The following subsections review their applications in software engineering.

\subsubsection{Automating Software Engineering Tasks}
Several studies have reviewed the opportunities and challenges of using LLMs in software engineering. \citet{hou2023large} provides a comprehensive analysis, categorizing LLMs into encoder-only, encoder-decoder, and decoder-only architectures. The increasing use of decoder-only models (e.g., GPT) for automating code generation and completion has significantly reduced manual effort. However, challenges persist, including handling domain-specific knowledge, improving dataset quality, and addressing the complexity of software engineering tasks beyond simple text generation. Key areas for improvement include better data preprocessing, fine-tuning for specific SE tasks, and incorporating diverse datasets, particularly from industrial contexts. Enhancing LLM robustness and refining evaluation metrics remain critical for real-world applications. While LLMs show promise in assisting SE, they are not yet capable of fully replacing human developers in complex processes.
\citet{phan2024hyperagent} presents HyperAgent, a multi-agent system designed for software engineering (SE) tasks using specialized agents like Planner, Navigator, Code Editor, and Executor. Evaluated on SWE-Bench and Defects4J, HyperAgent excels in GitHub issue resolution, repository-level code generation, and fault localization, outperforming specialized systems in complex multi-step tasks. It automates large-scale coding tasks (e.g., bug fixing, feature addition) using LLMs, reducing manual intervention. However, challenges include scalability across diverse environments and high computational costs. Future work aims to integrate version control, enhance explainability, and extend to specialized SE domains like security and performance optimization. \citet{xia2024agentless} introduces AGENTLESS, a streamlined approach automating SE tasks through a two-phase process: localization and repair. Unlike agent-based systems, AGENTLESS employs a hierarchical structure, reducing integration complexity while achieving 27.33\% on SWE-bench Lite. It automates bug localization and repair efficiently but struggles with cases lacking localization clues and complex reasoning tasks. Future improvements focus on expanding to more SE tasks, refining patch selection, and integrating testing frameworks, offering a simpler, cost-efficient alternative to multi-agent SE automation.

\subsubsection{Evaluating LLM for Software Development}
With the rise of generative AI, several studies have reviewed LLMs' feasibility in code generation and complex software development tasks.
\citet{liang2024can} examines GPT-4’s ability to replicate empirical SE research by generating analysis pipelines. While GPT-4 structured high-level plans well, it lacked domain-specific expertise, with only 30\% of its generated code executing without modification. Human oversight remains essential for ensuring accuracy. \citet{sandberg2024evaluating} evaluates GPT-4 in full-stack web development, highlighting its efficiency in generating functional applications for simple projects. However, as complexity increases, GPT-4 struggles with debugging and integration, requiring significant human intervention. \citet{gu2023effectiveness} assesses ChatGPT, CodeLlama, and PolyCoder in domain-specific coding, revealing struggles with API misuse. To address this, DomCoder integrates API recommendations and chain-of-thought prompting, improving domain-specific automation. However, challenges remain in sourcing domain-specific data and ensuring API consistency. 
\citet{fan2023large} reviews LLMs like GPT, BERT, and Codex in software engineering tasks, identifying strengths in code completion, bug detection, and automation. Challenges include hallucinations, non-deterministic outputs, and verification issues. Future directions involve improving prompt engineering, integrating LLMs with traditional SE methods, and enhancing automated testing to mitigate hallucinations and improve reliability.

\subsubsection{Code Generation}
Ongoing research continues to refine LLM-based code generation.
\citet{guo2024deepseek} introduces DeepSeek-Coder, an open-source alternative to proprietary models, with sizes ranging from 1.3B to 33B parameters trained on 2 trillion tokens across 87 programming languages. It enhances cross-file understanding using repository-level data construction and a Fill-in-the-Middle (FIM) approach, supporting a 16K context window for handling complex tasks. Benchmarks show it outperforms CodeLlama and StarCoder, even surpassing GPT-3.5 Turbo in some cases. With a permissive license, DeepSeek-Coder advances autonomous code generation and software development.
\citet{zhang2023planning} proposes Planning-Guided Transformer Decoding (PG-TD), which integrates a planning algorithm with Transformers to improve code generation by leveraging test case results. PG-TD surpasses traditional sampling and beam search, boosting pass rates on competitive programming benchmarks. However, it is computationally intensive and depends on existing test cases, limiting broader applications. Future work seeks to enhance efficiency through parallel tree search and automated test case generation for real-world software development.

In the GIS sector, \citet{hou2024can} highlights the limitations of general-purpose LLMs in geospatial code generation, citing issues such as coding hallucinations, deprecated functions, incompatible dependencies, and incorrect parameters. LLMs struggle with multi-modal geospatial data (e.g., raster, vector, elevation models) and fail to recognize built-in GIS functions in platforms like Google Earth Engine, OpenLayers, and Leaflet. Even advanced prompting provides only marginal improvements, exposing fundamental training gaps. To address these challenges, the study introduces GeoCode-Eval (GCE), an evaluation framework assessing LLMs' cognition, comprehension, and innovation in geospatial programming. Systematic testing reveals significant shortcomings across commercial and open-source models. The authors propose fine-tuning with domain-specific datasets, demonstrated by GEECode-GPT, a Code LLaMA-7B model trained on Google Earth Engine scripts, which significantly improves execution accuracy. Additionally, they introduce GeoCode-Bench, a benchmark dataset assessing LLM performance through multiple-choice, fill-in-the-blank, and coding tasks. Instruction-tuning with datasets from NASA Earth Data, USGS, and OpenStreetMap is also recommended to enhance geospatial reasoning. The study underscores LLMs' current limitations in geospatial programming and presents fine-tuning, benchmarking, and instruction-tuning as essential strategies to improve their reliability and usability for GIS applications.

\subsubsection{Front-end App Development from UI Prototypes}
\citet{xiao2024prototype2code} introduces Prototype2Code, an end-to-end framework for automating front-end code generation from UI design prototypes. Traditional UI-to-code methods often produce fragmented, unstructured code, impacting maintainability. Prototype2Code addresses this by integrating design linting, graph-based UI structuring, and LLMs to enhance code generation. It first detects and corrects UI inconsistencies through linting, constructs a hierarchical layout tree for structured components, and refines UI elements using a Graph Neural Network (GNN)-based classifier before generating modular HTML and CSS with LLMs. Benchmarks against CodeFun and GPT-4V-based Screenshot-to-Code show superior visual fidelity, readability, and maintainability, validated by SSIM, PSNR, and MSE metrics. A user study confirms reduced manual modifications and improved usability. Future work aims to support dynamic components, interactivity, and cross-platform adaptability. \citet{manuardi2024images} explores AI-driven front-end automation by converting UI mockups into structured code. Unlike text-based coding tools like GitHub Copilot, UI-driven development requires visual processing. The study proposes a multi-modal AI system that integrates computer vision and LLMs, using edge detection, contour analysis, and OCR to generate an intermediate representation, which a multi-modal LLM translates into front-end code for Angular and Bootstrap. Implemented at Blue Reply as a web-based tool, the system improves development efficiency by reducing manual coding while ensuring maintainability, advancing automated and intuitive front-end development.

\subsection{Limitation and Knowledge Gaps}
Despite advancements in LLM-powered code generation, challenges remain, particularly in scientific and GIS-based web applications. These limitations hinder seamless automation of front-end and back-end development, necessitating further research. The key challenges are outlined below as C1–C4.

\begin{description} \item[C1. Limited Graphical and Visual Prompting:] LLMs rely on text-based prompting, limiting their ability to interpret GUI designs and wireframe sketches. While image-to-code models show promise, they struggle with complex layouts, interactive elements, and contextual relationships. The lack of robust graph-based UI understanding restricts structured, modular front-end code generation.

\item[C2. Absence of Software Engineering Best Practices:] AI-generated code often lacks integration with industry design patterns (e.g., MVC, MVVM) \citep{xiao2024prototype2code}, leading to poor maintainability \citep{ghoshdesign, nguyen2023generative}. LLM-driven workflows rarely incorporate CI/CD pipelines, software testing, or version control, limiting their practical usability in large-scale development \citep{corona2025question, mendoza2024development}.

\item[C3. Lack of Domain Knowledge in GIS and Scientific Applications:] General-purpose LLMs struggle with GIS and scientific computing due to insufficient training on geospatial standards (GeoJSON, WMS, WFS), web mapping engines, and 3D data visualization \citep{zhang2024bb, mansourian2024chatgeoai, hou2024can}. While some fine-tuned models improve geospatial analysis \citep{hadid2024geoscience, hou2024geocode, akinboyewa2024gis}, web-based GIS dashboard generation remains largely unexplored.

\item[C4. Package Management Issues:] AI-generated code frequently suffers from dependency conflicts, outdated libraries, and compatibility issues, particularly in GIS and scientific computing \citep{hou2024can, mahmoudi2023development}. Web-based GIS applications require compatibility across Python, JavaScript, and C++ libraries (e.g., Django, Flask, OpenLayers, Leaflet), which LLMs often fail to handle effectively.

\end{description}

To bridge these gaps, a robust framework is needed to automate GIS web application development. This framework should allow users to input GUI sketches from non-technical tools (e.g., PowerPoint) for seamless, code-free development.

\section{Methodology}
\label{Methdology}
The following subsections outline our research motivation, the challenges we address, and our conceptual knowledge-driven approach. We then define the software design requirements, focusing on target users and key features. With these in place, we present the overall framework, detailing its major steps. This includes visual contextual prompting techniques, a software engineering and GIS knowledge base leveraging the RAG paradigm and vector databases, and Knowledge-Augmented Generation (KAG) for iterative code generation.

Rather than fine-tuning or training specific models, our approach emphasizes knowledge augmentation for LLM-driven coding. We propose a system-based, adaptable method that integrates software engineering principles into user-selected LLMs. Additionally, we introduce a generalizable approach that allows end-users to use custom UI wireframes as inputs, enabling LLMs to generate front-end web applications.

\subsection{Motivation and Contributions}
Our research develops a knowledge-driven approach to instruct LLMs, particularly GPTs, for automated code generation aligned with software engineering best practices. This supports the creation of robust, maintainable web-based GIS applications for scientific data sharing and visualization. By integrating industrial-grade practices and modern software stacks, our method lowers technical barriers for domain scientists with coding experience but limited software engineering expertise, enabling them to generate front-end CyberGIS applications efficiently.

To address the challenges in geospatial code generation identified in recent studies \citep{hou2024can, hou2024geocode}, particularly the limitations of general-purpose GPT models in handling built-in GIS and visualization libraries and managing complex package dependencies, we propose a novel knowledge-driven approach. Existing models frequently exhibit hallucinations, runtime errors, and an inability to generate complex CyberGIS applications with advanced data querying and visualization capabilities \citep{hou2024can}. Our approach seeks to overcome these limitations by integrating structured domain knowledge with intelligent prompt engineering and package management solutions, ensuring that LLM-generated code is accurate, executable, and aligned with industry standards. The key novelties of our contributions are as follows:
\begin{description}
\item[Knowledge-Augmented Code Generation:] Instead of fine-tuning LLMs, we employ Knowledge-Augmented Generation (KAG) to construct code-base and knowledge representations in a knowledge graph, integrating software engineering principles and real-world examples. This systematic approach improves the accuracy, maintainability, and scalability of LLM-generated GIS code.

\item[Visual Contextual Prompting:] Our Python-based system translates SVG-based wireframes into structured prompts, allowing non-programmers to design UI layouts in PowerPoint or Adobe Illustrator. The system extracts GUI components, spatial relationships, and functional annotations, querying a visual contextual knowledge base for implementation details compatible with React, Angular, Bootstrap, Tailwind CSS, Leaflet, and D3.js. This structured prompting method improves code accuracy, maintainability, and best-practice adherence, making LLM-assisted GIS development more accessible and reliable.

\item[Software Engineering Practices Integration:] We introduce a chain-of-steps guidance framework that ensures LLMs generate robust web-based GIS applications aligned with industry standards. Unlike conventional approaches, our framework enforces software design patterns such as MVVM in React-based SPAs, improving modularity and maintainability for GIS data visualization in web applications.

\item[Multi-tool Package Management:] LLM-generated code often suffers from package incompatibility and outdated dependencies. Instead of generating package management files directly, our method employs a structured knowledge-driven approach to guide industry-standard tools like NPM for JavaScript and TypeScript. This ensures efficient dependency resolution, reducing deployment failures and improving workflow reliability in GIS application development. 
\end{description}

This study presents the conceptual design and technical implementation of our framework, developed in Python within a Jupyter Notebook environment. Our research explores new paradigms to enhance LLM-driven code generation, focusing on flexibility and adaptability. Rather than tailoring to a specific LLM, our approach is a generalizable knowledge framework that integrates with various GPT-based models, ensuring scalability across AI-driven coding workflows.

\subsection{Design Requirements}
\label{subsec:DesignRequire}
The target users of our proposed framework are domain scientists with strong scientific computing and programming skills in geospatial analysis, simulations, and data modeling. However, they have limited or no exposure to software engineering principles and best practices for developing modular, maintainable front-end web applications using widely adopted software stacks and technologies in the IT industry.

The detailed technical design requirements of our proposed framework are as follows:
\begin{description}
    \item[R1. Support for UI Sketches and Wireframes as Inputs:]  
    The framework should accept UI sketches and wireframes as input, allowing users to visually define the structure and layout of their web applications without requiring extensive front-end development or cyberGIS expertise.
    
    \item[R2. Minimal UI/UX Expertise Required for Wireframing:]  
    Users should be able to create wireframes with minimal UI/UX design experience, ensuring accessibility for domain scientists without specialized front-end design skills. The framework should accept wireframes casually sketched using commonly available software, such as vector graphics created in Microsoft PowerPoint or Word, rather than requiring professional UI/UX prototyping tools like Adobe XD, Figma, or the discontinued Adobe Muse.
    
    \item[R3. Modular and Extensible Knowledge Base:]  
    The framework should provide a structured and expandable knowledge base that allows developers and open-source communities to incrementally refine and customize sample code. By integrating domain-specific and software engineering knowledge, the framework can enhance its ability to generate front-end code with advanced features while leveraging the latest code libraries—all while adhering to best practices.
    
    \item[R4. Integration of Software Engineering Practices:]  
    The system should incorporate established software design and architectural patterns to enhance code maintainability, scalability, and compliance with industry standards. For instance, modern web applications are rarely built using plain HTML, JavaScript, and a basic Python web server. As applications grow in complexity—integrating more features, data, and user workflows—it becomes essential to adopt software engineering best practices, such as Separation of Concerns (SoC) and the Singleton design pattern. These principles help modularize components, optimize performance, and promote code reusability, ultimately leading to the development of robust web frameworks that abstract these practices into MVC (Model-View-Controller) and MVVM (Model-View-ViewModel) architectures. We aim to develop a systematic approach to guide existing LLMs specialized in code generation for specific domain areas, ensuring they adhere to IT industry conventions and technologies (e.g., web frameworks like React) to generate robust and scalable front-end web applications.
    
    \item[R5. Multi-tool Paradigm for Package Management:]  
    A multi-tool approach should be employed to manage software dependencies, leveraging dedicated package management and versioning tools specific to each programming language (e.g., Conda for Python environments and NPM or NVM (Node Version Manager) for server-side JavaScript and TypeScript). This ensures robust package selection, version control, dependency management, and seamless integration of external libraries.
    
    \item[R6. Ease of Deployment and Execution:]  
    The generated applications should be easy to deploy and execute, minimizing the technical complexity required for setting up and running the resulting web-based GIS software. 
\end{description}
Based on these design requirements, we developed a comprehensive framework within a Python environment enabled through Jupyter Notebook. This framework enables the deployment of a code generation pipeline that produces a front-end codebase in a local file-based system. The choice of a Python Jupyter environment is justified by Python’s widespread use in GIS-related tasks, with many major spatial analysis libraries built in Python. More importantly, Python seamlessly integrates with APIs and platforms for both commercial and open-source LLMs, such as OpenAI's ChatGPT API and Ollama, a framework for running local LLMs.

\subsection{Framework Design}
\label{subsec:frameworkDesign}
\begin{figure*}[htbp]
 \centering
\includegraphics[width=\textwidth]{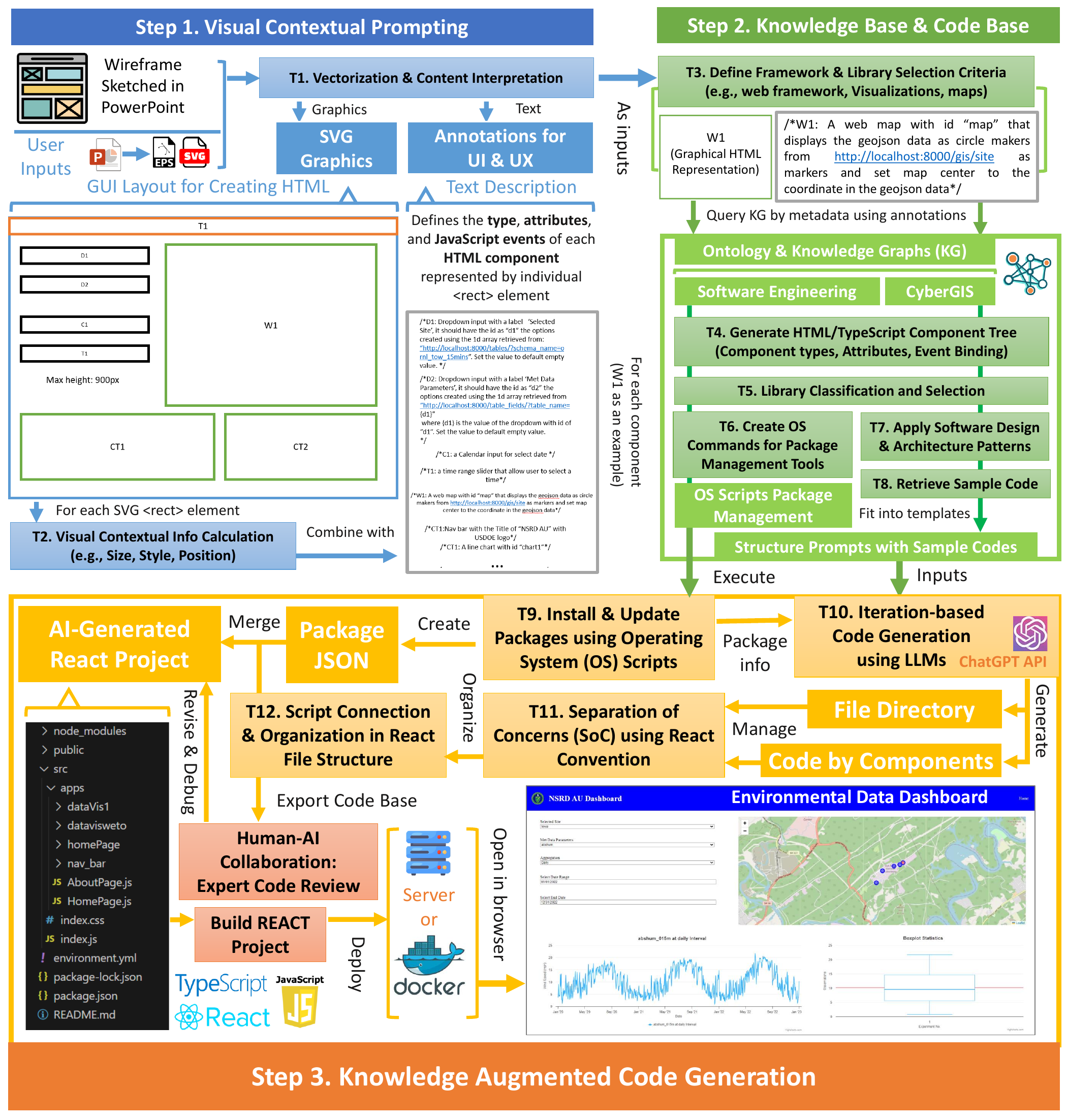}
\caption{Overall framework architecture for LLM-driven code generation using knowledge-augmented prompting and structured domain knowledge.}
 \label{fig:arch}
\end{figure*}

The overall framework design is illustrated in Figure \ref{fig:arch}, where we define 12 tasks (T1–T12) categorized into three major steps that form the core of the code generation workflow. These steps include: (1) Visual Contextual Prompting, (2) Knowledge Base and Code Base, and (3) Knowledge-Augmented Code Generation. The framework takes as input one or more user-defined wireframes that specify the layout and features of the GUI across different pages. As output, it generates the complete code base for a React project.

\subsubsection{Visual Contextual Prompting}
\label{subsec:simulations1}
We introduce a novel visual contextual prompting technique, implemented using Python scripts, to translate GUI wireframes into structured prompts for guiding LLMs in generating high-quality JavaScript and TypeScript code. Unlike conventional text-based prompting, our approach enables non-technical users, such as domain scientists, to design web applications visually using PowerPoint and Adobe Illustrator.

Existing graphical prompting methods often rely on computer vision (CV)-based techniques to interpret bitmap images (e.g., PNG, JPEG). In contrast, our method processes SVG-based wireframes, which offer greater editability and flexibility during prototyping. This structured approach allows users to define UI components without programming expertise.

This step consists of two tasks, T1 and T2, aligned with the design requirements R1 and R2 in Subsection \ref{subsec:DesignRequire}.

\begin{description}
\item[T1. Vectorization \& Content Interpretation]
A Python program processes wireframe files exported from PowerPoint or Adobe Illustrator in EPS/SVG formats, extracting spatial and contextual information to construct HTML layouts and components. Each GUI element—such as dropdown menus, charts, and web maps—is represented as a vector graphical entity with embedded annotations specifying its function (e.g., data visualization, mapping interface, UI control). These annotations also capture interactions and dependencies, guiding LLMs in event binding for dynamic UI behaviors.

\item[T2. Visual Contextual Info Calculation]
Another Python script calculates visual contextual information for each SVG element, mapping it to an HTML component. This includes position, size, and style, which are then combined with wireframe annotations to generate engineered prompts. These prompts define not only individual components but also layout structures, interdependencies, and event handling logic, ensuring that LLM-generated code aligns with industry standards and preserves the user’s design intent.
\end{description}
By integrating visual design with structured knowledge-based prompting, our technique lowers the barrier for non-programmers, enabling them to develop web-based GIS and data visualization applications without direct coding expertise.

\subsubsection{Knowledge Base and Code Base}
\label{subsec:KB_CB}
The input information, including graphical components, visual context, and annotations from user, defined wireframes—captures GUI design and software requirements in plain, non-technical language. This enables users without coding expertise to define application requirements effortlessly. However, a combined knowledge base and code base is essential to interpret these context-aware visual prompts, transforming them into structured software development strategies. These strategies generate enriched prompts with sample code, guiding LLMs for code generation within a RAG framework.

Our knowledge base is built using knowledge graphs that encapsulate software development experiences, system requirements, and architectural designs from previous projects on digital twins, cyber infrastructure, and web-based visual analytics dashboards, as shown in Table \ref{tab:KG_previousProjects}. These knowledge graphs document and classify software stacks, componentization methods, and system designs based on domain-specific use cases and data types, ensuring structured and informed LLM-assisted code generation.

\begin{scriptsize}
\begin{table*}[h]
\footnotesize
\centering
\caption{Knowledge Base Construction from Previous Research Projects} \label{tab:KG_previousProjects} 
\begin{tabular*}{\textwidth}{|p{3.2 cm}|p{3cm}|p{3cm}|p{4.5cm}|}
 \hline
\multicolumn{1}{c}{\textbf{Application Type}}  
& \multicolumn{1}{c}{\textbf{Knowledge Domain Classification}} 
& \multicolumn{1}{c}{\textbf{Software Stacks}} 
& \multicolumn{1}{c}{\textbf{References}}  \\
\hline
 Urban Digital Twins & Intelligent Transportation, GIS, Data Visualization  & Angular, Openlayers, WebGL, D3JS & \citet{li2024empowering, xu2021continuous, xu2023smart, niloofar2023general} \\ 
 \hline
 Environment Data Dashboards & Hydroinformatics, GIS & Google Map APIs, Leaflet, High-chart, React & \citet{xu2019web2, xu2022overview, xu2020web, xu2019web} \\
 \hline
 Immersive Reality Apps & 3D City Modeling & ThreeJS, WebGL, Unity 3D Game Engine & \citet{xu2023toward, xu2024semi} \\
 \hline
 Visual Analytics Dashboards & Transportation, Power Grid, Data Visualization, Visual Analytics & Leaflet, D3JS, High-chart & \citet{chinthavali2022alternative, berres2021explorative, berres2021multiscale, xu2022interactive, xu2022geo, xu2021visualizing, shao2022computer, xu2024explainable, muste2017community}  \\
 \hline
\end{tabular*}
\end{table*}
\end{scriptsize}

\begin{figure*}[htbp]
 \centering
\includegraphics[width=\textwidth]{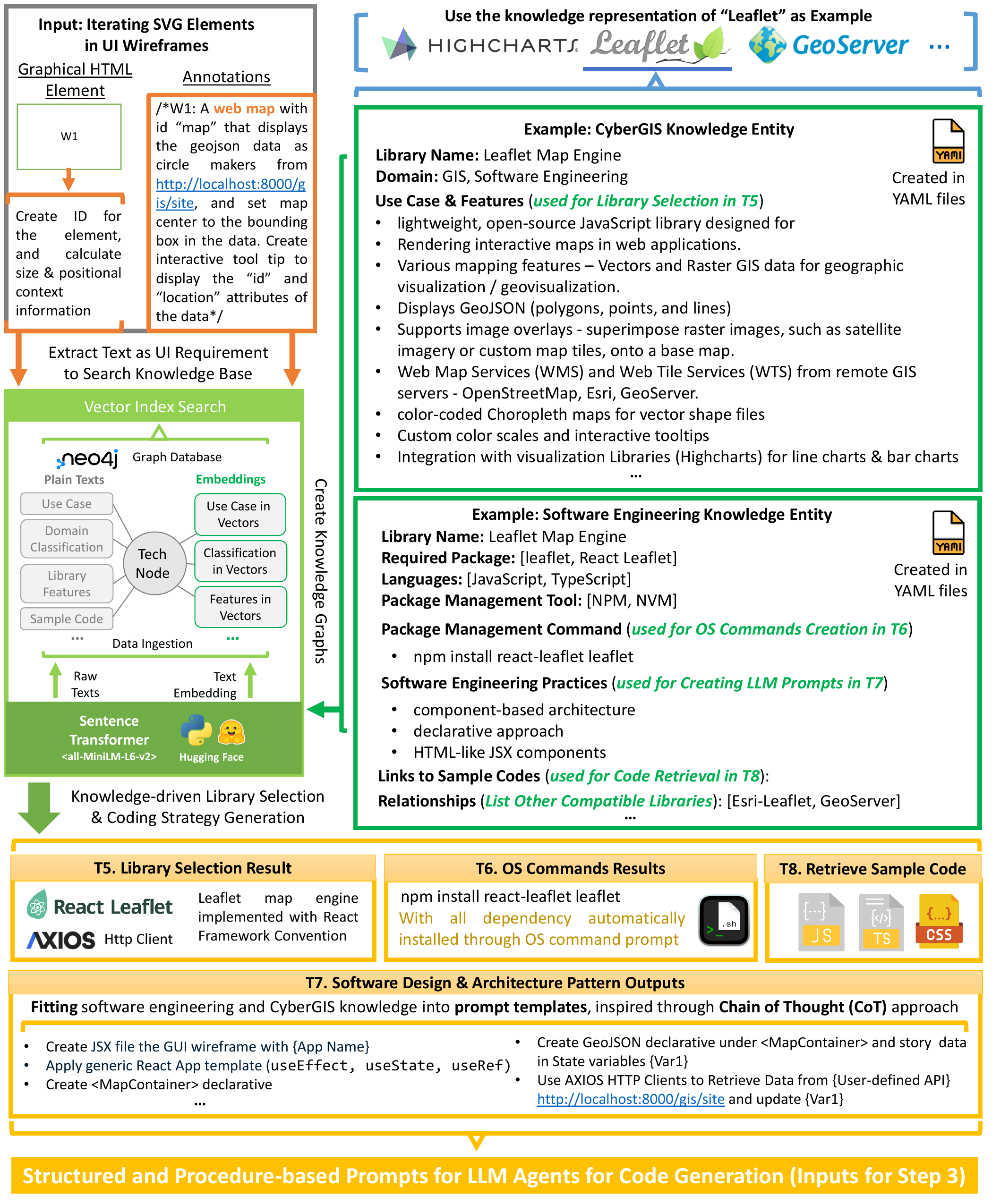}
\caption{Structured knowledge representations for converting plain-language annotations from UI wireframes into structured prompts with technical terminology.}
 \label{fig:KG_method}
\end{figure*}

An example of the knowledge graph structure is depicted in Figure \ref{fig:KG_method}. The sample code stored within these graphs is either excerpted and refined from previous projects or augmented using the ChatGPT API under expert review and supervision.

The process of knowledge-driven prompting follows a procedural approach, structured through tasks T3 to T8, ensuring a systematic and context-driven software development workflow. These tasks are detailed in the following list, with their rationale depicted in Figure \ref{fig:KG_method}.  

\begin{description}
\item[T3. Framework \& Library Selection]
This task defines criteria for selecting web frameworks and libraries based on wireframe annotations. A vector index search on a knowledge graph (covering software engineering and CyberGIS domains) maps UI elements (e.g., date selector, GIS map) to React components (e.g., calendar dropdown, web map engine). The knowledge graph stores metadata on HTML elements, events, and best practices, ensuring context-aware, industry-aligned component selection.

\item[T4. HTML/TypeScript Component Tree Generation]
Using extracted GUI components and annotations, this task constructs a structured HTML/TypeScript component tree, preserving semantic correctness, hierarchical relationships, and CSS styles. The wireframe context guides nested structures and component interactions to ensure logical front-end design.

\item[T5. Front-End Library Selection]
Based on T3's criteria, this task selects appropriate front-end software stacks, including CSS frameworks, UI libraries, mapping tools, and data visualization frameworks. It determines whether to use standard HTML elements or third-party solutions (e.g., Material-UI, Tailwind CSS, D3.js, Leaflet) by performing a vector index search on a Neo4j-backed knowledge graph \citep{xu2024automating}, ensuring optimal usability, performance, and maintainability.

\item[T6. OS Commands for Package Management]
After selecting libraries in T5, a Python script retrieves the corresponding OS commands for installing, updating, and managing dependencies using appropriate package management tools. JavaScript/TypeScript dependencies use NPM/NVM, Python libraries use Pip/Conda, and system-level packages use APT (Debian) or Yum (RHEL). The script consolidates commands into .sh or .bash scripts, enabling one-click execution for package management across Windows and Linux DevOps environments, handling installation, version control, and conflict resolution.

\item[T7. Software Design \& Architecture Patterns]
This task ensures adherence to scalable and modular software design principles by integrating Separation of Concerns (SoC), MVC, and MVVM patterns. It leverages industry-standard frameworks (React, Angular) and structured prompt templates to guide LLMs in generating componentized code with proper event binding and state management.

\item[T8. Software Development Prompt Generation]
Expanding on T6, this task refines structured prompts for LLM-driven code generation, incorporating component-based development, lazy loading, and dynamic rendering. Sample code from the knowledge base is embedded into prompts, providing step-by-step procedural instructions to ensure AI-generated code adheres to modular design principles for reusability, maintainability, and readability.
\end{description}

Our knowledge base and code repository serve as domain experts in software engineering and GIS, facilitating the translation of plain-language annotations from wireframes into structured software development strategies. By leveraging technical terminology and software engineering best practices, we provide system-based instructions with sample code, enabling AI agents to generate robust and well-structured code in the next step.

\subsubsection{Knowledge Augmented Code Generation}
\label{subsec:simulations3}
\begin{figure*}[htbp]
 \centering
\includegraphics[width=\textwidth]{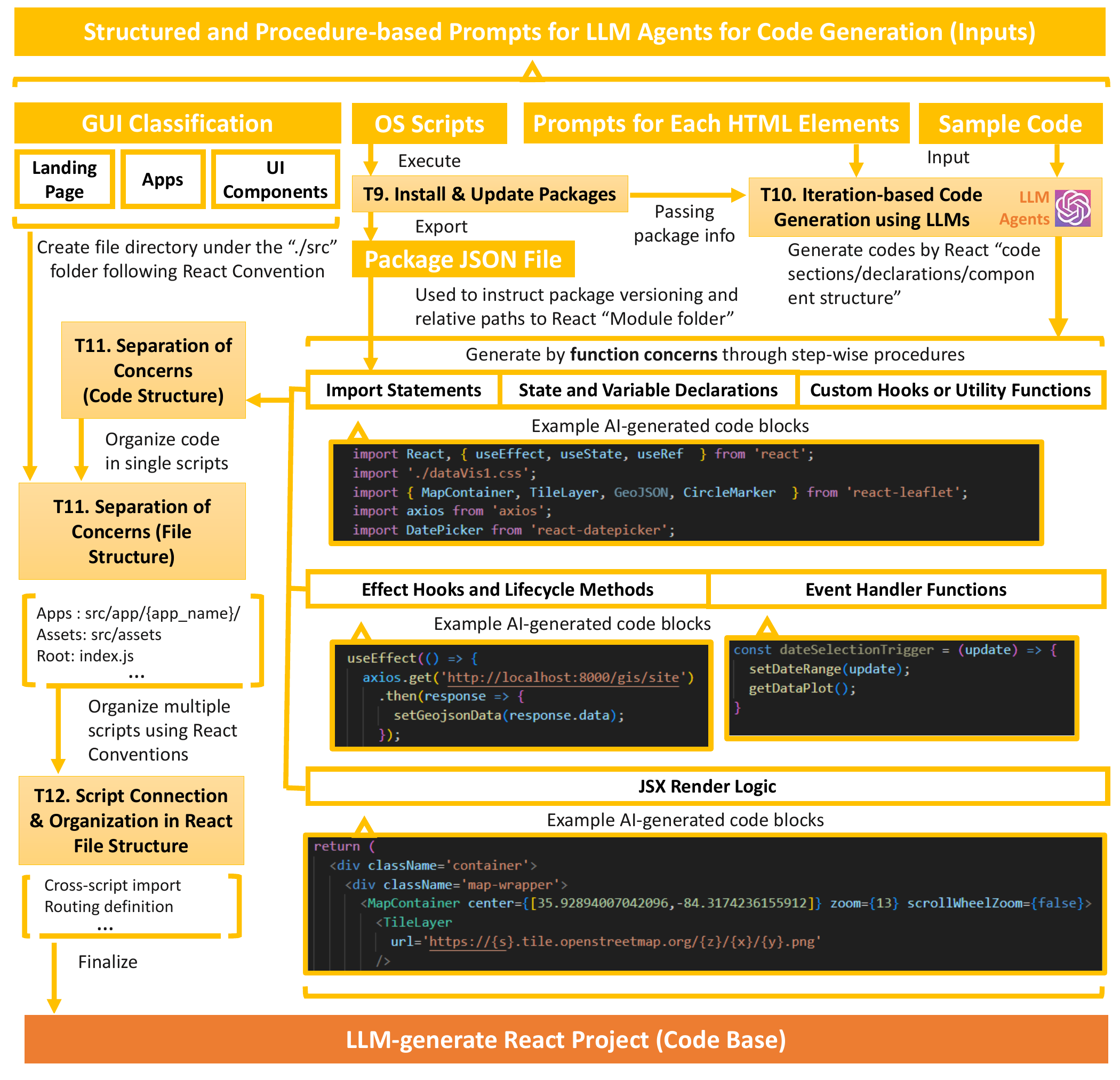}
 \caption{A procedure-based, knowledge-augmented code generation process that aligns with the conventions and best practices of the React framework. }
 \label{fig:kag}
\end{figure*}

The structured prompts generated from T8 are then fed into the LLM agent to enable knowledge-augmented code generation. Given the token limit constraints of LLMs, we adopt a procedure-based, iterative approach that combines rule-based Python scripting with the customized generation capabilities of LLMs. This method ensures efficient front-end code generation using the React framework, which follows the Model-View-ViewModel (MVVM) architecture, along with established software engineering conventions and React’s standard file structure. Our KAG approach entails the following tasks from T9 to T11, which are illustrated in Figure \ref{fig:kag}. 

\begin{description}
    \item[T9. Install \& Update Packages using Operating System (OS) Scripts]  
    This task automates the installation and updating of required dependencies using OS-level scripts. It ensures that all necessary packages, including Node.js, npm, and front-end libraries (e.g., React, Material-UI, Leaflet), are properly installed and version-controlled. The scripts also handle package updates to maintain compatibility with evolving software frameworks and dependencies.

    \item[T10. Iteration-based Code Generation using LLMs]  
    Given the token limitations of LLMs, this task adopts a procedure-based, iterative approach for code generation. Rule-based Python scripts and structured prompts guide the LLMs to generate React components in a stepwise manner, ensuring compliance with best practices in componentization, event handling, and state management. The iterative process allows refinement and optimization of the generated code, minimizing redundancy and improving maintainability.

    \item[T11. Separation of Concerns (SoC) using React Convention]  
    This task enforces Separation of Concerns (SoC) by structuring the generated code according to React’s MVVM architecture. The React components, logic, styles, and API handlers are modularized into separate files, ensuring that UI elements, business logic, and data handling remain distinct. This improves scalability, maintainability, and code readability while aligning with industry standards.

    \item[T12. Script Connection \& Organization in React File Structure]  
    This task organizes and connects the generated scripts following the **React file structure convention**. The components are structured within a modular directory system (e.g., `components/`, `hooks/`, `services/`, `contexts/`). The script integration ensures that React Router, Redux (if applicable), and event-driven logic** are properly linked, allowing seamless front-end development with maintainable and reusable components.
\end{description}

After generating a complete React project through knowledge-augmented LLMs, two additional steps are required before the DevOps stage to ensure functionality, maintainability, and deployment readiness:

\begin{description}
\item[Expert Code Review: Human-AI Collaboration for Code Refinement]
A Human-AI collaboration process refines and validates the AI-generated React code. Human experts review, debug, and revise the code to ensure correctness, performance, and adherence to industry standards. Since our method follows React’s MVVM architecture and Separation of Concerns (SoC) principles, the AI-generated code is well-structured, readable, and maintainable, making it easier for developers to understand, modify, and integrate into production.

\item[Build React Project: Executing OS Scripts and Dependency Installation]
Once validated, the React project must be built and configured for execution. Users install dependencies by running the OS scripts generated in T9 via the command line in Windows or Linux. This ensures that React, Material-UI, Leaflet, and other libraries are correctly installed and up to date. After setup, the project is built using NPM or NPX commands, ensuring a fully functional front-end dashboard.
\end{description}
 
For deployment, the AI-generated React project can be hosted on a server (monolithic architecture) or deployed in a Docker container (microservice architecture). This ensures scalability, portability, and interoperability with databases and server-side APIs. By adopting this approach, AI-generated front-end components seamlessly integrate into enterprise systems, enhancing efficiency, maintainability, and modularity.

\section{Result Demonstration}
\label{subsec:Data}
Using our prototyping framework, we automated the generation of front-end code for two web-based GIS dashboards addressing distinct use cases involving environmental and energy infrastructure data. To showcase the advantages of the proposed framework, which leverages knowledge-augmented code generation guided by software engineering best practices and industry standards, we developed both dashboards as single-page applications using the React framework, integrating them within the same React project. 

\subsection{Case Study I - Meteorological Data Dashboard}
Access to continuous and high-quality meteorological data is essential for understanding regional climatology and atmospheric processes. Such data plays a crucial role in research efforts focused on assessing local climate patterns, modeling atmospheric dispersion, evaluating emissions, and ensuring environmental and operational safety. Research institutions like Oak Ridge National Laboratory (ORNL) require reliable meteorological measurements to support site operation, emergency preparedness, and environmental monitoring. However, meteorological data collection is often subject to various challenges, including sensor degradation, power fluctuations, lightning strikes, and instrument failures, all of which can introduce uncertainties and affect data reliability \citep{steckler_2025}. 

To address these challenges, this study aims to leverage the prototyping framework's ability to generate the code base for a robust visual dashboard to enhances the quality and usability of meteorological data collected at ORNL by placing domain experts in the loop to supervise the data collocation and quality control processes. Specifically, a comprehensive quality assessment was conducted using a statistical framework to process of five years of meteorological data, ensuring data integrity and continuity. The visual dashboard is developed to assist the visual exploration and supervision of the data outputs from the statistical framework 
The primary objective is to produce a high-quality, gap-filled dataset that supports accurate atmospheric dispersion modeling, which is crucial for understanding pollutant dispersion and regional air quality dynamics.

The study focuses on meteorological data collected from the Oak Ridge Reservation (ORR), located in East Tennessee. ORR is characterized by a complex ridge-and-valley topography, which significantly influences local wind patterns and atmospheric dispersion processes. The region’s terrain-driven microclimate poses challenges for meteorological modeling, making it essential to have high-quality, site-specific meteorological data. ORNL operates on-site meteorological towers designed to capture critical weather variables, providing a valuable resource for climatological and atmospheric studies.

\begin{figure*}[htbp]
 \centering
\includegraphics[width=\textwidth]{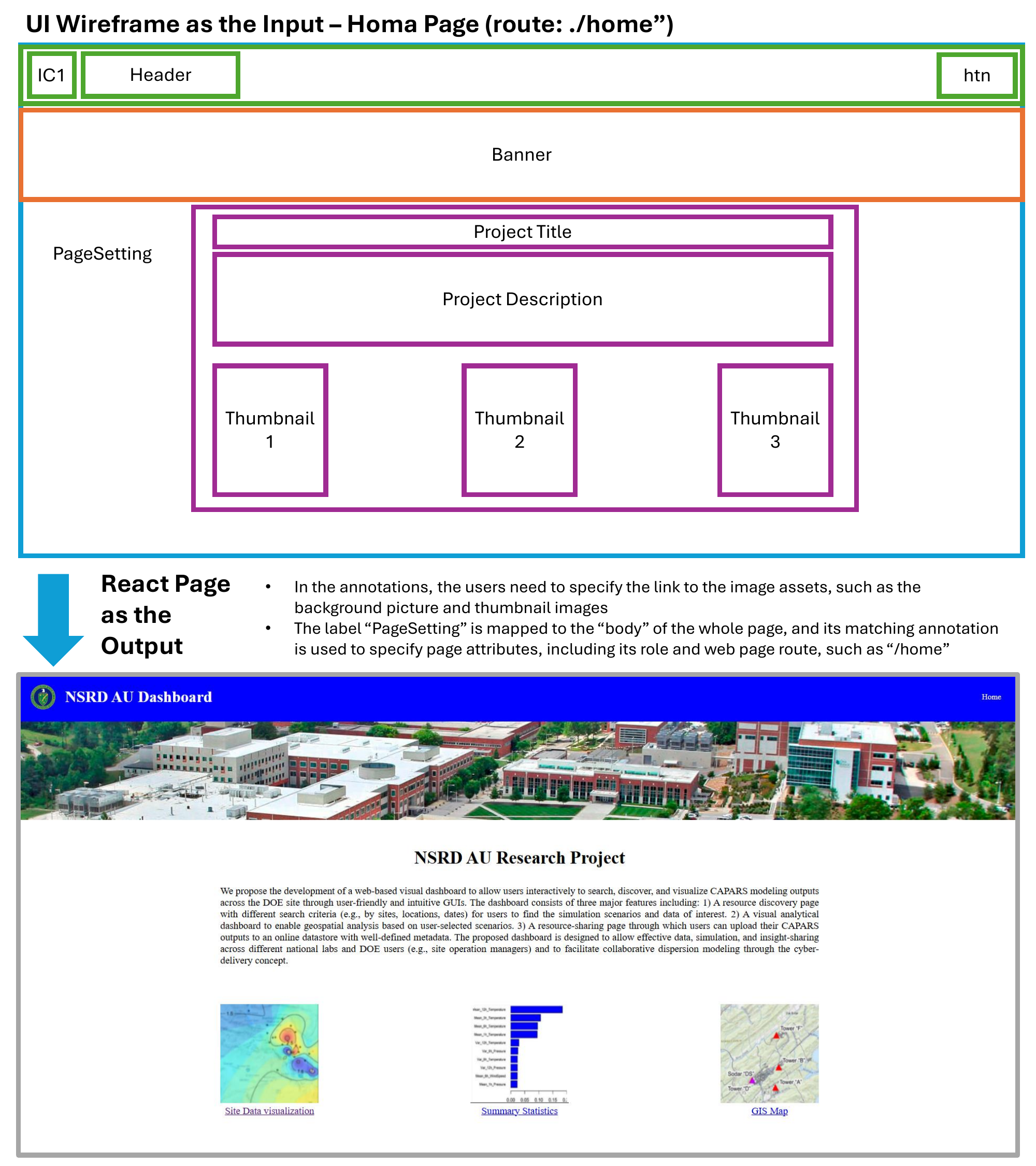}
 \caption{The UI wireframe for the homepage of the web-based application includes thumbnails that serve as navigation links, directing users to the dashboard generated for Use Case I. }
 \label{fig:demo-1}
\end{figure*}

\begin{figure*}[htbp]
 \centering
\includegraphics[width=\textwidth]{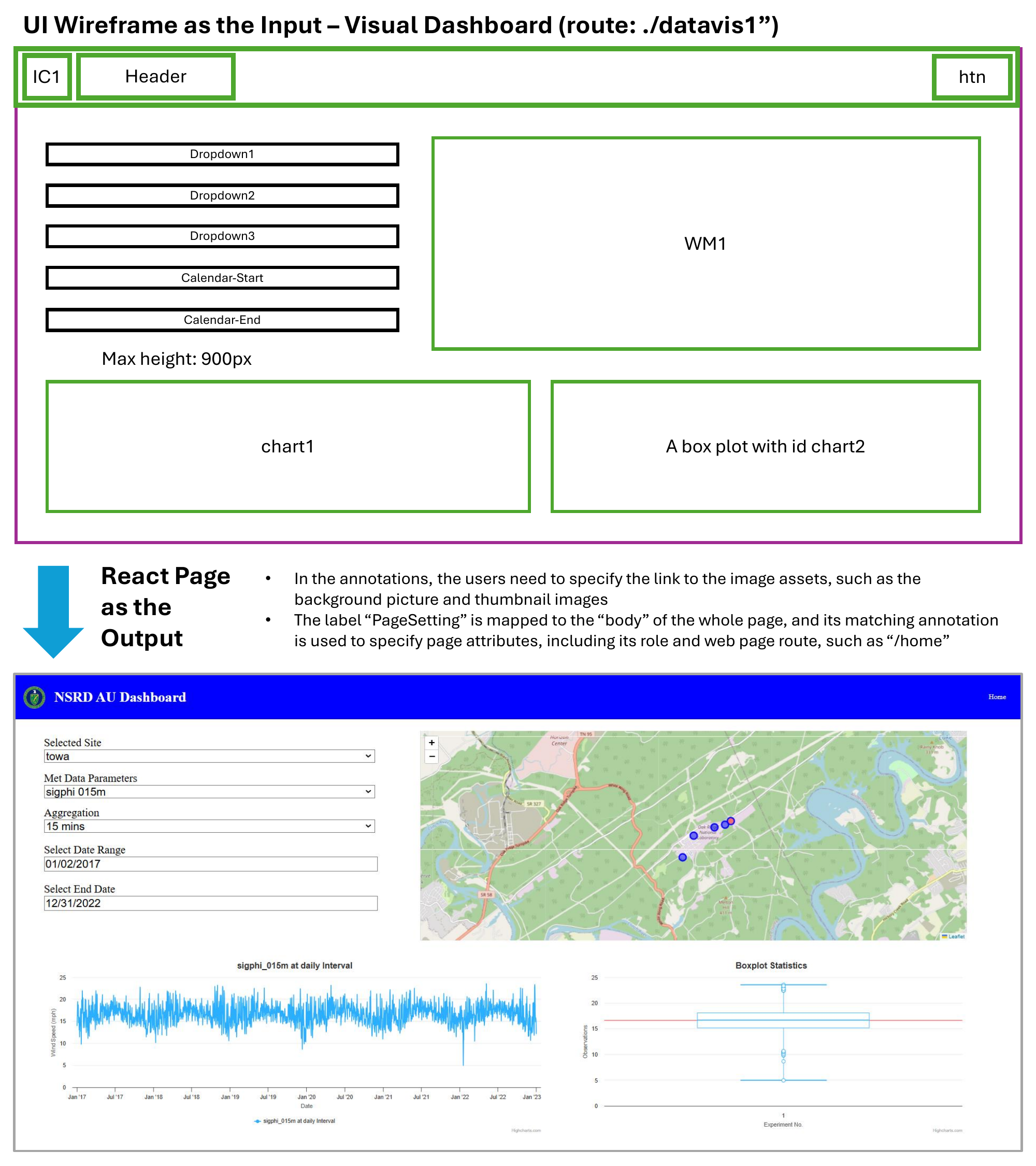}
 \caption{The UI wireframe for the visual dashboard of Use Case I is designed for visualizing meteorological data, incorporating time-series data from tower sensors and shapefiles representing site locations. }
 \label{fig:demo-2}
\end{figure*}

\subsection{Case Study II - Wind Turbine and Landuse Data}
The expansion of renewable energy infrastructure, such as wind farms, has raised concerns about its potential ecological impacts on bird habitats. Previous studies have assessed bird habitats using bird-watching surveys and remote sensing data on natural vegetation cover, offering valuable insights into avian ecology. Building on these methods, this case study investigates the hypothesis that noise and land cover changes resulting from wind turbine operations may displace grassland- and forest-dwelling birds. To explore this, we conducted a preliminary study using data from the United States Geological Survey (USGS)’s Wind Turbine Database (USWTDB) and correlated it with a 20-year time series of land cover changes from the Multi-Resolution Land Characteristics (MRLC)’s National Land Cover Database (NLCD).

This case study focuses on wind farm sites across multiple states in the United States. The USWTDB provides detailed GIS data on wind turbine locations, construction years, and operational specifications, which are linked to land cover changes documented by the NLCD. The NLCD dataset includes high-resolution raster-based land cover data, capturing variations in vegetation and natural land cover over two decades. By comparing land cover data before and after the establishment of wind farms, the study identifies patterns of vegetation loss and fragmentation caused by infrastructure development, including roads, facilities, and pavements.

To validate these findings, this case study aims to develop a web-based GIS dashboard that integrates the time series of land cover changes from the NLCD dataset with wind turbine locations retrieved from the USWTDB. The dashboard visually overlays land cover rasters with wind turbine locations, enabling users to assess potential land cover changes caused by wind farm operations. This tool highlights areas where wildlife conservation strategies may be needed. By providing insights into the ecological impacts of wind farms, this approach establishes a practical framework for mitigating habitat loss and protecting avian species affected by renewable energy development.

\begin{figure*}[htbp]
 \centering
\includegraphics[width=\textwidth]{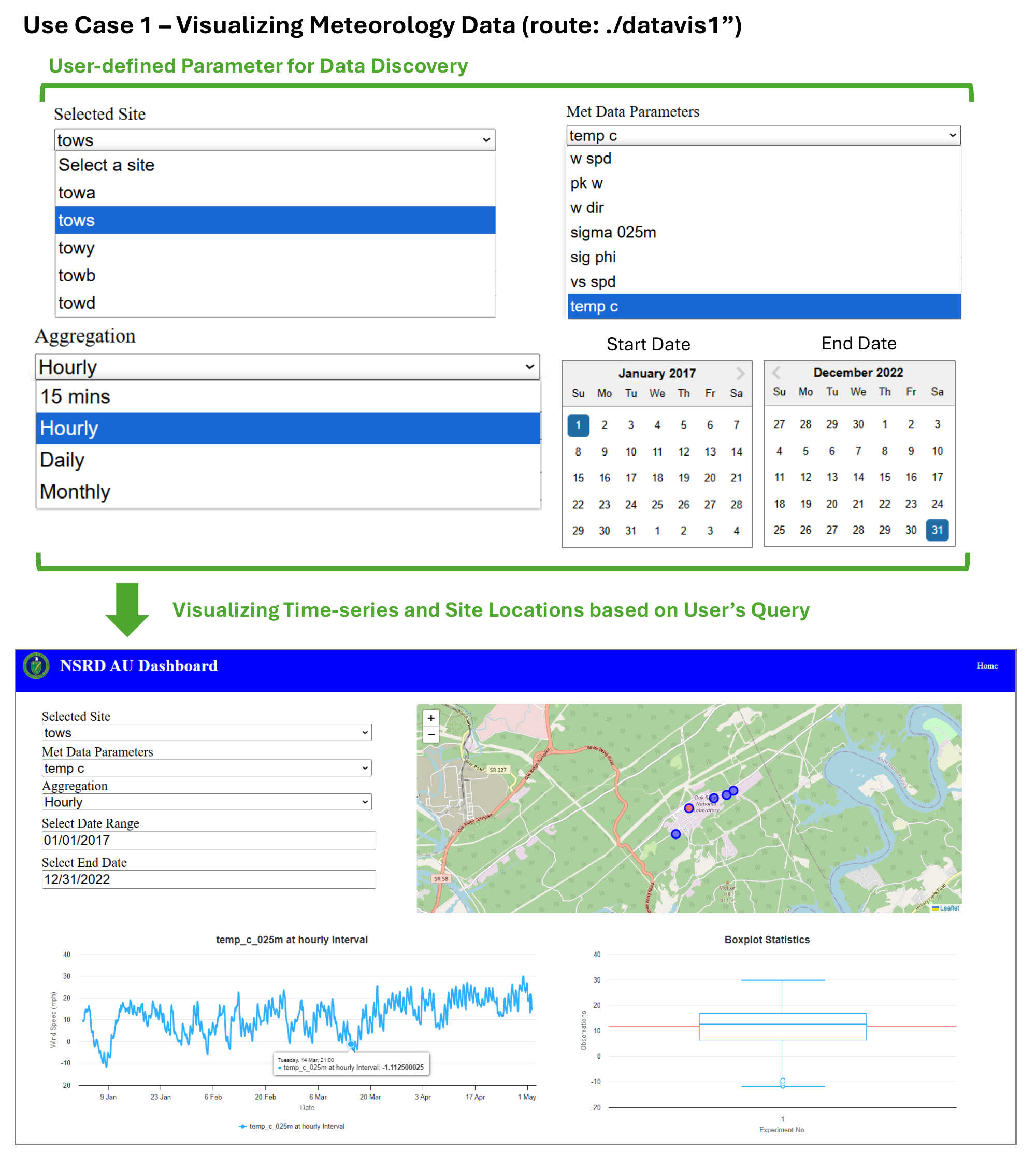}
 \caption{A demonstration of the dashboard's capability to visualize meteorological data (e.g., shapfiles and time-series) for Use Case I. }
 \label{fig:demo-3}
\end{figure*}

\begin{figure*}[htbp]
 \centering
\includegraphics[width=\textwidth]{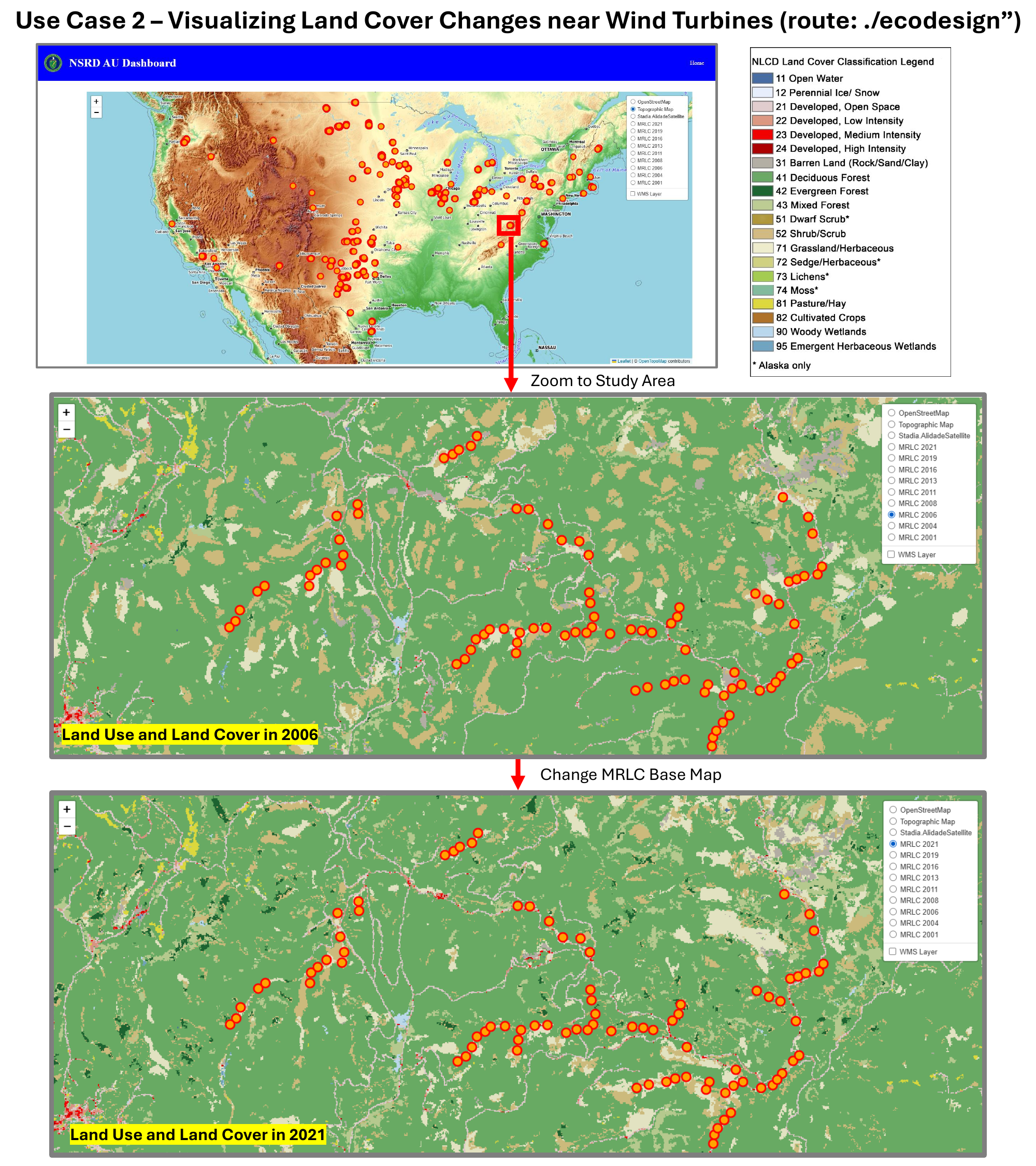}
 \caption{A demonstration of the dashboard's capability to visualize land use and land cover raster data for Use Case II.}
 \label{fig:demo-4}
\end{figure*}

\subsection{Context-aware Visual Prompting}
We present the UIs of the web-based application, which are generated from user-defined wireframes for the homepage (as depicted in Figure \ref{fig:demo-1}) and the visual dashboard for Use Case I (as illustrated in Figure \ref{fig:demo-2}). These UIs are integrated as distinct routes within a single React project, enhancing scalability and extensibility to accommodate additional GIS dashboards under the same project while maintaining a consistent UI style through code reuse. This approach improves code maintainability by adhering to the singleton software engineering principle, implemented using the React framework.

The context-aware visual prompting technique is designed exclusively for generating structured instructions and prompts to enable the LLM to produce front-end code. In our case studies, we developed the back-end application separately to provide web-based API endpoints for querying and retrieving data from the database using the Python FastAPI framework. However, the development of the back-end application is not within the scope of our proposed framework.

\subsection{AI-generated Dashboard Demonstration}
For the meteorological data dashboard, the AI-generated interface provides an interactive platform for exploring time-series meteorological data collected from the ORR. The dashboard enables real-time visualization of meteorological variables captured by tower sensors, including temperature, wind speed, humidity, and atmospheric pressure (as shown in Figure \ref{fig:demo-3}). The AI-driven dashboard generation process ensures that data integrity is preserved by integrating statistical quality assessments to identify missing or inconsistent measurements.

Users can interact with the dashboard to:
\begin{enumerate}
\item Query sensor measurements at different sites, with their locations visualized on the map.
\item Visualize time-series meteorological trends for a large number of parameters over different time periods.
\item Display statistical summaries of the selected time-series data.
\end{enumerate}
The AI-generated dashboard effectively places domain experts in the loop, allowing them to supervise data quality and validate automated statistical assessments, thereby improving the usability of long-term meteorological data sets for atmospheric modeling and environmental monitoring.

For Use Case II, the AI-generated dashboard integrates GIS-based land cover data with wind turbine locations, providing a comprehensive platform for analyzing the ecological impacts of wind energy infrastructure (as depicted in Figure \ref{fig:demo-4}). By utilizing AI-assisted dashboard generation, the system automatically organizes spatial raster data sets and overlays them with wind turbine distributions to reveal patterns of land cover transformation.
The dashboard enables users to perform the following functionalities:
\begin{enumerate}
    \item Examine land cover changes before and after wind farm construction using a 20-year historical data set. Compare the extent of vegetation loss and landscape fragmentation near wind turbine sites.
    \item Assess potential ecological risks associated with wind farm expansion by correlating turbine operations with habitat shifts.
\end{enumerate}
This AI-enhanced approach to dashboard generation significantly reduces development time while improving the consistency and maintainability of GIS visualization tools. By combining AI-assisted UI design with data-driven analysis, this study demonstrates the potential of AI in advancing interactive environmental monitoring systems.

\subsection{Limitation and Future Work}
As our study primarily focuses on prototyping a knowledge-driven framework to demonstrate the feasibility of guiding LLMs for adaptive, pattern-driven code generation in creating robust web-based GIS applications using software engineering best practices and industry-grade web frameworks, we do not delve into theoretical or algorithmic research for evaluating the performance of various LLMs and generative AI technologies. Instead, the primary emphasis is on the application and use cases presented in the study. However, our existing framework has several limitations, as listed below: 

\begin{description}
    \item[Lack of Comparative Analysis across Different LLMs] The framework's performance and effectiveness were evaluated using a single large language model (LLM) to demonstrate feasibility. Future work should include comparative studies across various LLMs to assess their suitability for different domains and coding scenarios.
    
    \item[Requires Human Expert Review of AI-Generated Code] While the framework automates front-end code generation, the resulting code still requires human experts to review for correctness, optimization, and adherence to specific project requirements. Future advancements could integrate automatic validation tools or explainable AI mechanisms to reduce dependency on manual reviews.

    \item[Customized Functions Require Manual Programming Efforts] Although the framework automates standard functionalities, developing highly customized features still requires manual programming, which limits full automation. Future iterations of the framework could incorporate a mechanism to better support user-defined customizations through enhanced prompt engineering or plug-and-play modular components.

    \item[Limited Support for Backend Integration] The current framework focuses on front-end code generation with minimal backend integration capabilities. Extending the framework to support full-stack development workflows, including database and API integration, would be a valuable addition.

    \item[Scalability to Larger Projects] The prototyping framework has been demonstrated on two dashboards within a single React project. Future work could explore its scalability to larger and more complex multi-application systems, addressing performance and maintainability challenges.

    \item[Generalizability to Non-React Frameworks] The framework is currently optimized for React-based projects. Future research should evaluate its adaptability to other popular front-end frameworks, such as Angular or Vue.js, to enhance its applicability across diverse development environments.
\end{description}

It is anticipated that the next stage efforts will include conducting a comparative analysis of multiple LLMs to evaluate their performance and adaptability in generating front-end code for GIS web applications. Automated validation mechanisms should also be developed to ensure correctness, optimization, and adherence to best practices, reducing reliance on human review. Additionally, enhancing the framework's support for user-defined customizations would streamline the development of complex features without significant manual effort. Expanding the framework to include backend development capabilities, such as API and database integration, would enable full-stack automation. To broaden applicability, the framework should be generalized to support multiple front-end frameworks, including Angular and Vue.js. Real-time collaboration features can further improve the framework by enabling teams to seamlessly work on AI-generated code. Lastly, domain-specific enhancements tailored to areas like environmental monitoring and urban planning could optimize the framework for specialized use cases.
\section{Conclusion}
\label{Conclusion}
This study presents a knowledge-augmented code generation framework that integrates domain expertise, software engineering principles, and Generative AI to automate GIS-based web application development. Leveraging Context-Aware Visual Prompting and RAG, it transforms user-defined UI wireframes into scalable, maintainable front-end code for environmental and energy data visualization dashboards. The framework bridges the gap between domain scientists and software engineering, enabling users with minimal web development experience to generate functional GIS applications. Case studies demonstrate its effectiveness in generating two interactive dashboards: a meteorological data dashboard for visualizing time-series and spatial datasets from tower sensors, and a wind turbine and land use dashboard overlaying wind farm locations with land cover change data. These AI-generated dashboards support real-time exploration and analysis, ensuring data integrity and usability for scientific research and policy-making. By adopting a modular React-based architecture and integrating software engineering best practices such as MVVM and Separation of Concerns (SoC), the framework enhances scalability, maintainability, and reusability. It highlights the potential of AI-driven UI generation to reduce development time, improve consistency, and streamline GIS visualization tool creation.

Despite these advancements, the framework currently focuses on front-end code generation, requiring manual backend integration. Human expert review remains necessary for validating AI-generated code, optimizing performance, and ensuring compliance with project requirements. Future research will explore automated validation, backend integration, and support for additional front-end frameworks like Angular and Vue.js. Comparative evaluations of LLMs will assess their effectiveness in scientific software development. This work advances AI-assisted software development by pioneering a structured, knowledge-driven approach to LLM-powered code generation. By enhancing accessibility and automation in GIS web development, it lays the foundation for AI-driven environmental monitoring, digital twins, and smart city analytics.
\bibliographystyle{elsarticle-harv}
 \bibliography{bib_file}

\begin{thebibliography}{78}
\expandafter\ifx\csname natexlab\endcsname\relax\def\natexlab#1{#1}\fi
\providecommand{\url}[1]{\texttt{#1}}
\providecommand{\href}[2]{#2}
\providecommand{\path}[1]{#1}
\providecommand{\DOIprefix}{doi:}
\providecommand{\ArXivprefix}{arXiv:}
\providecommand{\URLprefix}{URL: }
\providecommand{\Pubmedprefix}{pmid:}
\providecommand{\doi}[1]{\href{http://dx.doi.org/#1}{\path{#1}}}
\providecommand{\Pubmed}[1]{\href{pmid:#1}{\path{#1}}}
\providecommand{\bibinfo}[2]{#2}
\ifx\xfnm\relax \def\xfnm[#1]{\unskip,\space#1}\fi
\bibitem[{Akinboyewa et~al.(2024)Akinboyewa, Li, Ning and
  Lessani}]{akinboyewa2024gis}
\bibinfo{author}{Akinboyewa, T.}, \bibinfo{author}{Li, Z.},
  \bibinfo{author}{Ning, H.}, \bibinfo{author}{Lessani, M.N.},
  \bibinfo{year}{2024}.
\newblock \bibinfo{title}{Gis copilot: Towards an autonomous gis agent for
  spatial analysis}.
\newblock \bibinfo{journal}{arXiv preprint arXiv:2411.03205} .
\bibitem[{Alatalo et~al.(2017)Alatalo, Pouke, Koskela, Hurskainen, Florea and
  Ojala}]{alatalo2017two}
\bibinfo{author}{Alatalo, T.}, \bibinfo{author}{Pouke, M.},
  \bibinfo{author}{Koskela, T.}, \bibinfo{author}{Hurskainen, T.},
  \bibinfo{author}{Florea, C.}, \bibinfo{author}{Ojala, T.},
  \bibinfo{year}{2017}.
\newblock \bibinfo{title}{Two real-world case studies on 3d web applications
  for participatory urban planning}, in: \bibinfo{booktitle}{Proceedings of the
  22nd International Conference on 3D Web Technology}, pp.
  \bibinfo{pages}{1--9}.
\bibitem[{Baldazzi et~al.(2023)Baldazzi, Bellomarini, Ceri, Colombo, Gentili
  and Sallinger}]{baldazzi2023fine}
\bibinfo{author}{Baldazzi, T.}, \bibinfo{author}{Bellomarini, L.},
  \bibinfo{author}{Ceri, S.}, \bibinfo{author}{Colombo, A.},
  \bibinfo{author}{Gentili, A.}, \bibinfo{author}{Sallinger, E.},
  \bibinfo{year}{2023}.
\newblock \bibinfo{title}{{Fine-tuning large enterprise language models via
  ontological reasoning}}, in: \bibinfo{booktitle}{International Joint
  Conference on Rules and Reasoning}, \bibinfo{organization}{Springer}. pp.
  \bibinfo{pages}{86--94}.
\bibitem[{Berres et~al.(2021a)Berres, LaClair, Wang, Xu, Ravulaparthy, Todd,
  Tennille and Sanyal}]{berres2021multiscale}
\bibinfo{author}{Berres, A.S.}, \bibinfo{author}{LaClair, T.J.},
  \bibinfo{author}{Wang, C.}, \bibinfo{author}{Xu, H.},
  \bibinfo{author}{Ravulaparthy, S.}, \bibinfo{author}{Todd, A.},
  \bibinfo{author}{Tennille, S.A.}, \bibinfo{author}{Sanyal, J.},
  \bibinfo{year}{2021}a.
\newblock \bibinfo{title}{Multiscale and multivariate transportation system
  visualization for shopping district traffic and regional traffic}.
\newblock \bibinfo{journal}{Transportation Research Record}
  \bibinfo{volume}{2675}, \bibinfo{pages}{23--37}.
\bibitem[{Berres et~al.(2021b)Berres, Xu, Tennille, Severino, Ravulaparthy and
  Sanyal}]{berres2021explorative}
\bibinfo{author}{Berres, A.S.}, \bibinfo{author}{Xu, H.},
  \bibinfo{author}{Tennille, S.A.}, \bibinfo{author}{Severino, J.},
  \bibinfo{author}{Ravulaparthy, S.}, \bibinfo{author}{Sanyal, J.},
  \bibinfo{year}{2021}b.
\newblock \bibinfo{title}{Explorative visualization for traffic safety using
  adaptive study areas}.
\newblock \bibinfo{journal}{Transportation research record}
  \bibinfo{volume}{2675}, \bibinfo{pages}{51--69}.
\bibitem[{Brown et~al.(2020)Brown, Mann, Ryder, Subbiah, Kaplan, Dhariwal,
  Neelakantan, Shyam, Sastry, Askell et~al.}]{brown2020language}
\bibinfo{author}{Brown, T.}, \bibinfo{author}{Mann, B.},
  \bibinfo{author}{Ryder, N.}, \bibinfo{author}{Subbiah, M.},
  \bibinfo{author}{Kaplan, J.D.}, \bibinfo{author}{Dhariwal, P.},
  \bibinfo{author}{Neelakantan, A.}, \bibinfo{author}{Shyam, P.},
  \bibinfo{author}{Sastry, G.}, \bibinfo{author}{Askell, A.}, et~al.,
  \bibinfo{year}{2020}.
\newblock \bibinfo{title}{{Language models are few-shot learners}}.
\newblock \bibinfo{journal}{Advances in neural information processing systems}
  \bibinfo{volume}{33}, \bibinfo{pages}{1877--1901}.
\bibitem[{Chinthavali et~al.(2022)Chinthavali, Hasan, Yoginath, Xu, Nugent,
  Jones, Engebretsen, Olatt, Tansakul, Christopher
  et~al.}]{chinthavali2022alternative}
\bibinfo{author}{Chinthavali, S.}, \bibinfo{author}{Hasan, S.S.},
  \bibinfo{author}{Yoginath, S.}, \bibinfo{author}{Xu, H.},
  \bibinfo{author}{Nugent, P.}, \bibinfo{author}{Jones, T.},
  \bibinfo{author}{Engebretsen, C.}, \bibinfo{author}{Olatt, J.},
  \bibinfo{author}{Tansakul, V.}, \bibinfo{author}{Christopher, C.}, et~al.,
  \bibinfo{year}{2022}.
\newblock \bibinfo{title}{An alternative timing and synchronization approach
  for situational awareness and predictive analytics}, in:
  \bibinfo{booktitle}{2022 IEEE 23rd International Conference on Information
  Reuse and Integration for Data Science (IRI)}, \bibinfo{organization}{IEEE}.
  pp. \bibinfo{pages}{172--177}.
\bibitem[{Corona-Fraga et~al.(2025)Corona-Fraga, Hernandez-Suarez,
  Sanchez-Perez, Toscano-Medina, Perez-Meana, Portillo-Portillo,
  Olivares-Mercado and Garc{\'\i}a~Villalba}]{corona2025question}
\bibinfo{author}{Corona-Fraga, P.}, \bibinfo{author}{Hernandez-Suarez, A.},
  \bibinfo{author}{Sanchez-Perez, G.}, \bibinfo{author}{Toscano-Medina, L.K.},
  \bibinfo{author}{Perez-Meana, H.}, \bibinfo{author}{Portillo-Portillo, J.},
  \bibinfo{author}{Olivares-Mercado, J.},
  \bibinfo{author}{Garc{\'\i}a~Villalba, L.J.}, \bibinfo{year}{2025}.
\newblock \bibinfo{title}{Question--answer methodology for vulnerable source
  code review via prototype-based model-agnostic meta-learning}.
\newblock \bibinfo{journal}{Future Internet} \bibinfo{volume}{17},
  \bibinfo{pages}{33}.
\bibitem[{Dembski et~al.(2020)Dembski, W{\"o}ssner, Letzgus, Ruddat and
  Yamu}]{dembski2020urban}
\bibinfo{author}{Dembski, F.}, \bibinfo{author}{W{\"o}ssner, U.},
  \bibinfo{author}{Letzgus, M.}, \bibinfo{author}{Ruddat, M.},
  \bibinfo{author}{Yamu, C.}, \bibinfo{year}{2020}.
\newblock \bibinfo{title}{Urban digital twins for smart cities and citizens:
  The case study of herrenberg, germany}.
\newblock \bibinfo{journal}{Sustainability} \bibinfo{volume}{12},
  \bibinfo{pages}{2307}.
\bibitem[{Fan et~al.(2023)Fan, Gokkaya, Harman, Lyubarskiy, Sengupta, Yoo and
  Zhang}]{fan2023large}
\bibinfo{author}{Fan, A.}, \bibinfo{author}{Gokkaya, B.},
  \bibinfo{author}{Harman, M.}, \bibinfo{author}{Lyubarskiy, M.},
  \bibinfo{author}{Sengupta, S.}, \bibinfo{author}{Yoo, S.},
  \bibinfo{author}{Zhang, J.M.}, \bibinfo{year}{2023}.
\newblock \bibinfo{title}{Large language models for software engineering:
  Survey and open problems}, in: \bibinfo{booktitle}{2023 IEEE/ACM
  International Conference on Software Engineering: Future of Software
  Engineering (ICSE-FoSE)}, \bibinfo{organization}{IEEE}. pp.
  \bibinfo{pages}{31--53}.
\bibitem[{Fayad et~al.(2015)Fayad, Sanchez, Hegde, Basia and
  Vakil}]{fayad2015software}
\bibinfo{author}{Fayad, M.}, \bibinfo{author}{Sanchez, H.A.},
  \bibinfo{author}{Hegde, S.G.}, \bibinfo{author}{Basia, A.},
  \bibinfo{author}{Vakil, A.}, \bibinfo{year}{2015}.
\newblock \bibinfo{title}{Software patterns, knowledge maps, and domain
  analysis}.
\newblock \bibinfo{publisher}{CRC Press}.
\bibitem[{Ferr{\'e}-Bigorra et~al.(2022)Ferr{\'e}-Bigorra, Casals and
  Gangolells}]{ferre2022adoption}
\bibinfo{author}{Ferr{\'e}-Bigorra, J.}, \bibinfo{author}{Casals, M.},
  \bibinfo{author}{Gangolells, M.}, \bibinfo{year}{2022}.
\newblock \bibinfo{title}{The adoption of urban digital twins}.
\newblock \bibinfo{journal}{Cities} \bibinfo{volume}{131},
  \bibinfo{pages}{103905}.
\bibitem[{Garg et~al.(2018)Garg, Aryal, Wang, Shah, Kecskemeti and
  Ranjan}]{garg2018cloud}
\bibinfo{author}{Garg, S.}, \bibinfo{author}{Aryal, J.}, \bibinfo{author}{Wang,
  H.}, \bibinfo{author}{Shah, T.}, \bibinfo{author}{Kecskemeti, G.},
  \bibinfo{author}{Ranjan, R.}, \bibinfo{year}{2018}.
\newblock \bibinfo{title}{Cloud computing based bushfire prediction for
  cyber--physical emergency applications}.
\newblock \bibinfo{journal}{Future Generation Computer Systems}
  \bibinfo{volume}{79}, \bibinfo{pages}{354--363}.
\bibitem[{Ghosh and Team(2024)}]{ghoshdesign}
\bibinfo{author}{Ghosh, D.P.}, \bibinfo{author}{Team, D.A.},
  \bibinfo{year}{2024}.
\newblock \bibinfo{title}{Design orchestration of conveying system} .
\bibitem[{Ghosh et~al.(2017)Ghosh, Pragathi, Ullas and
  Borra}]{ghosh2017intelligent}
\bibinfo{author}{Ghosh, R.}, \bibinfo{author}{Pragathi, R.},
  \bibinfo{author}{Ullas, S.}, \bibinfo{author}{Borra, S.},
  \bibinfo{year}{2017}.
\newblock \bibinfo{title}{Intelligent transportation systems: A survey}, in:
  \bibinfo{booktitle}{2017 International Conference on Circuits, Controls, and
  Communications (CCUBE)}, \bibinfo{organization}{IEEE}. pp.
  \bibinfo{pages}{160--165}.
\bibitem[{Gu et~al.(2023)Gu, Chen, Lin, Hu, Zhang, Wan, Wei, Xu and
  Wang}]{gu2023effectiveness}
\bibinfo{author}{Gu, X.}, \bibinfo{author}{Chen, M.}, \bibinfo{author}{Lin,
  Y.}, \bibinfo{author}{Hu, Y.}, \bibinfo{author}{Zhang, H.},
  \bibinfo{author}{Wan, C.}, \bibinfo{author}{Wei, Z.}, \bibinfo{author}{Xu,
  Y.}, \bibinfo{author}{Wang, J.}, \bibinfo{year}{2023}.
\newblock \bibinfo{title}{On the effectiveness of large language models in
  domain-specific code generation}.
\newblock \bibinfo{journal}{ACM Transactions on Software Engineering and
  Methodology} .
\bibitem[{Guo et~al.(2024)Guo, Zhu, Yang, Xie, Dong, Zhang, Chen, Bi, Wu, Li
  et~al.}]{guo2024deepseek}
\bibinfo{author}{Guo, D.}, \bibinfo{author}{Zhu, Q.}, \bibinfo{author}{Yang,
  D.}, \bibinfo{author}{Xie, Z.}, \bibinfo{author}{Dong, K.},
  \bibinfo{author}{Zhang, W.}, \bibinfo{author}{Chen, G.}, \bibinfo{author}{Bi,
  X.}, \bibinfo{author}{Wu, Y.}, \bibinfo{author}{Li, Y.}, et~al.,
  \bibinfo{year}{2024}.
\newblock \bibinfo{title}{Deepseek-coder: When the large language model meets
  programming--the rise of code intelligence}.
\newblock \bibinfo{journal}{arXiv preprint arXiv:2401.14196} .
\bibitem[{Hadid et~al.(2024)Hadid, Chakraborty and Busby}]{hadid2024geoscience}
\bibinfo{author}{Hadid, A.}, \bibinfo{author}{Chakraborty, T.},
  \bibinfo{author}{Busby, D.}, \bibinfo{year}{2024}.
\newblock \bibinfo{title}{When geoscience meets generative ai and large
  language models: Foundations, trends, and future challenges}.
\newblock \bibinfo{journal}{Expert Systems} , \bibinfo{pages}{e13654}.
\bibitem[{Hou et~al.(2024a)Hou, Shen, Liang, Zhao, Gui, Li and Wu}]{hou2024can}
\bibinfo{author}{Hou, S.}, \bibinfo{author}{Shen, Z.}, \bibinfo{author}{Liang,
  J.}, \bibinfo{author}{Zhao, A.}, \bibinfo{author}{Gui, Z.},
  \bibinfo{author}{Li, R.}, \bibinfo{author}{Wu, H.}, \bibinfo{year}{2024}a.
\newblock \bibinfo{title}{Can large language models generate geospatial code?}
\newblock \bibinfo{journal}{arXiv preprint arXiv:2410.09738} .
\bibitem[{Hou et~al.(2024b)Hou, Shen, Zhao, Liang, Gui, Guan, Li and
  Wu}]{hou2024geocode}
\bibinfo{author}{Hou, S.}, \bibinfo{author}{Shen, Z.}, \bibinfo{author}{Zhao,
  A.}, \bibinfo{author}{Liang, J.}, \bibinfo{author}{Gui, Z.},
  \bibinfo{author}{Guan, X.}, \bibinfo{author}{Li, R.}, \bibinfo{author}{Wu,
  H.}, \bibinfo{year}{2024}b.
\newblock \bibinfo{title}{Geocode-gpt: A large language model for geospatial
  code generation tasks}.
\newblock \bibinfo{journal}{arXiv preprint arXiv:2410.17031} .
\bibitem[{Hou et~al.(2023)Hou, Zhao, Liu, Yang, Wang, Li, Luo, Lo, Grundy and
  Wang}]{hou2023large}
\bibinfo{author}{Hou, X.}, \bibinfo{author}{Zhao, Y.}, \bibinfo{author}{Liu,
  Y.}, \bibinfo{author}{Yang, Z.}, \bibinfo{author}{Wang, K.},
  \bibinfo{author}{Li, L.}, \bibinfo{author}{Luo, X.}, \bibinfo{author}{Lo,
  D.}, \bibinfo{author}{Grundy, J.}, \bibinfo{author}{Wang, H.},
  \bibinfo{year}{2023}.
\newblock \bibinfo{title}{Large language models for software engineering: A
  systematic literature review}.
\newblock \bibinfo{journal}{ACM Transactions on Software Engineering and
  Methodology} .
\bibitem[{Ikegwu et~al.(2022)Ikegwu, Nweke, Anikwe, Alo and
  Okonkwo}]{ikegwu2022big}
\bibinfo{author}{Ikegwu, A.C.}, \bibinfo{author}{Nweke, H.F.},
  \bibinfo{author}{Anikwe, C.V.}, \bibinfo{author}{Alo, U.R.},
  \bibinfo{author}{Okonkwo, O.R.}, \bibinfo{year}{2022}.
\newblock \bibinfo{title}{Big data analytics for data-driven industry: a review
  of data sources, tools, challenges, solutions, and research directions}.
\newblock \bibinfo{journal}{Cluster Computing} \bibinfo{volume}{25},
  \bibinfo{pages}{3343--3387}.
\bibitem[{Jia et~al.(2019)Jia, Komeily, Wang and Srinivasan}]{jia2019adopting}
\bibinfo{author}{Jia, M.}, \bibinfo{author}{Komeily, A.},
  \bibinfo{author}{Wang, Y.}, \bibinfo{author}{Srinivasan, R.S.},
  \bibinfo{year}{2019}.
\newblock \bibinfo{title}{Adopting internet of things for the development of
  smart buildings: A review of enabling technologies and applications}.
\newblock \bibinfo{journal}{Automation in Construction} \bibinfo{volume}{101},
  \bibinfo{pages}{111--126}.
\bibitem[{Kadupitige(2022)}]{kadupitige2022enhancing}
\bibinfo{author}{Kadupitige, J.C.S.K.}, \bibinfo{year}{2022}.
\newblock \bibinfo{title}{Enhancing Molecular Dynamics Simulations with Machine
  Learning and Advanced Cyberinfrastructure}.
\newblock \bibinfo{publisher}{Indiana University}.
\bibitem[{Kampmann et~al.(2019)Kampmann, Alrifaee, Kohout, W{\"u}stenberg,
  Woopen, Nolte, Eckstein and Kowalewski}]{kampmann2019dynamic}
\bibinfo{author}{Kampmann, A.}, \bibinfo{author}{Alrifaee, B.},
  \bibinfo{author}{Kohout, M.}, \bibinfo{author}{W{\"u}stenberg, A.},
  \bibinfo{author}{Woopen, T.}, \bibinfo{author}{Nolte, M.},
  \bibinfo{author}{Eckstein, L.}, \bibinfo{author}{Kowalewski, S.},
  \bibinfo{year}{2019}.
\newblock \bibinfo{title}{A dynamic service-oriented software architecture for
  highly automated vehicles}, in: \bibinfo{booktitle}{2019 IEEE Intelligent
  Transportation Systems Conference (ITSC)}, \bibinfo{organization}{IEEE}. pp.
  \bibinfo{pages}{2101--2108}.
\bibitem[{Kim et~al.(2022)Kim, Yoon, Lee, Mago, Lee and Cho}]{kim2022design}
\bibinfo{author}{Kim, D.}, \bibinfo{author}{Yoon, Y.}, \bibinfo{author}{Lee,
  J.}, \bibinfo{author}{Mago, P.J.}, \bibinfo{author}{Lee, K.},
  \bibinfo{author}{Cho, H.}, \bibinfo{year}{2022}.
\newblock \bibinfo{title}{Design and implementation of smart buildings: A
  review of current research trend}.
\newblock \bibinfo{journal}{Energies} \bibinfo{volume}{15},
  \bibinfo{pages}{4278}.
\bibitem[{Kim et~al.(2017)Kim, Zimmermann, DeLine and Begel}]{kim2017data}
\bibinfo{author}{Kim, M.}, \bibinfo{author}{Zimmermann, T.},
  \bibinfo{author}{DeLine, R.}, \bibinfo{author}{Begel, A.},
  \bibinfo{year}{2017}.
\newblock \bibinfo{title}{Data scientists in software teams: State of the art
  and challenges}.
\newblock \bibinfo{journal}{IEEE Transactions on Software Engineering}
  \bibinfo{volume}{44}, \bibinfo{pages}{1024--1038}.
\bibitem[{Lei et~al.(2023)Lei, Janssen, Stoter and
  Biljecki}]{lei2023challenges}
\bibinfo{author}{Lei, B.}, \bibinfo{author}{Janssen, P.},
  \bibinfo{author}{Stoter, J.}, \bibinfo{author}{Biljecki, F.},
  \bibinfo{year}{2023}.
\newblock \bibinfo{title}{Challenges of urban digital twins: A systematic
  review and a delphi expert survey}.
\newblock \bibinfo{journal}{Automation in Construction} \bibinfo{volume}{147},
  \bibinfo{pages}{104716}.
\bibitem[{Li et~al.(2022)Li, Zhang, Liu, Qian and Zhang}]{li2022bibliometric}
\bibinfo{author}{Li, R.}, \bibinfo{author}{Zhang, H.}, \bibinfo{author}{Liu,
  C.}, \bibinfo{author}{Qian, Z.C.}, \bibinfo{author}{Zhang, L.},
  \bibinfo{year}{2022}.
\newblock \bibinfo{title}{Bibliometric and visualized analysis of user
  experience design research: From 1999 to 2019}.
\newblock \bibinfo{journal}{Sage Open} \bibinfo{volume}{12},
  \bibinfo{pages}{21582440221087266}.
\bibitem[{Li et~al.(2021)Li, Xu, Huang, Guo, Kang and Ye}]{li2021emerging}
\bibinfo{author}{Li, X.}, \bibinfo{author}{Xu, H.}, \bibinfo{author}{Huang,
  X.}, \bibinfo{author}{Guo, C.}, \bibinfo{author}{Kang, Y.},
  \bibinfo{author}{Ye, X.}, \bibinfo{year}{2021}.
\newblock \bibinfo{title}{Emerging geo-data sources to reveal human mobility
  dynamics during covid-19 pandemic: Opportunities and challenges}.
\newblock \bibinfo{journal}{Computational Urban Science} \bibinfo{volume}{1},
  \bibinfo{pages}{1--9}.
\bibitem[{Li et~al.(2024)Li, Xu, Tupayachi, Omitaomu and
  Wang}]{li2024empowering}
\bibinfo{author}{Li, X.}, \bibinfo{author}{Xu, H.}, \bibinfo{author}{Tupayachi,
  J.}, \bibinfo{author}{Omitaomu, O.}, \bibinfo{author}{Wang, X.},
  \bibinfo{year}{2024}.
\newblock \bibinfo{title}{Empowering cognitive digital twins with generative
  foundation models: Developing a low-carbon integrated freight transportation
  system}.
\newblock \bibinfo{journal}{arXiv preprint arXiv:2410.18089} .
\bibitem[{Li and Ning(2023)}]{li2023autonomous}
\bibinfo{author}{Li, Z.}, \bibinfo{author}{Ning, H.}, \bibinfo{year}{2023}.
\newblock \bibinfo{title}{{Autonomous GIS: the next-generation AI-powered
  GIS}}.
\newblock \bibinfo{journal}{International Journal of Digital Earth}
  \bibinfo{volume}{16}, \bibinfo{pages}{4668--4686}.
\bibitem[{Liang et~al.(2024)Liang, Badea, Bird, DeLine, Ford, Forsgren and
  Zimmermann}]{liang2024can}
\bibinfo{author}{Liang, J.T.}, \bibinfo{author}{Badea, C.},
  \bibinfo{author}{Bird, C.}, \bibinfo{author}{DeLine, R.},
  \bibinfo{author}{Ford, D.}, \bibinfo{author}{Forsgren, N.},
  \bibinfo{author}{Zimmermann, T.}, \bibinfo{year}{2024}.
\newblock \bibinfo{title}{Can gpt-4 replicate empirical software engineering
  research?}
\newblock \bibinfo{journal}{Proceedings of the ACM on Software Engineering}
  \bibinfo{volume}{1}, \bibinfo{pages}{1330--1353}.
\bibitem[{Liu et~al.(2015)Liu, Padmanabhan and Wang}]{liu2015cybergis}
\bibinfo{author}{Liu, Y.}, \bibinfo{author}{Padmanabhan, A.},
  \bibinfo{author}{Wang, S.}, \bibinfo{year}{2015}.
\newblock \bibinfo{title}{Cybergis gateway for enabling data-rich geospatial
  research and education}.
\newblock \bibinfo{journal}{Concurrency and Computation: Practice and
  Experience} \bibinfo{volume}{27}, \bibinfo{pages}{395--407}.
\bibitem[{Liukko et~al.(2024)Liukko, Knappe, Anttila, Hakala, Ketola, Lahtinen,
  Poranen, Ritala, Set{\"a}l{\"a}, H{\"a}m{\"a}l{\"a}inen
  et~al.}]{liukko2024chatgpt}
\bibinfo{author}{Liukko, V.}, \bibinfo{author}{Knappe, A.},
  \bibinfo{author}{Anttila, T.}, \bibinfo{author}{Hakala, J.},
  \bibinfo{author}{Ketola, J.}, \bibinfo{author}{Lahtinen, D.},
  \bibinfo{author}{Poranen, T.}, \bibinfo{author}{Ritala, T.M.},
  \bibinfo{author}{Set{\"a}l{\"a}, M.},
  \bibinfo{author}{H{\"a}m{\"a}l{\"a}inen, H.}, et~al., \bibinfo{year}{2024}.
\newblock \bibinfo{title}{Chatgpt as a full-stack web developer}, in:
  \bibinfo{booktitle}{Generative AI for Effective Software Development}.
  \bibinfo{publisher}{Springer}, pp. \bibinfo{pages}{197--215}.
\bibitem[{Mahmoudi et~al.(2023)Mahmoudi, Camboim and
  Brovelli}]{mahmoudi2023development}
\bibinfo{author}{Mahmoudi, H.}, \bibinfo{author}{Camboim, S.},
  \bibinfo{author}{Brovelli, M.A.}, \bibinfo{year}{2023}.
\newblock \bibinfo{title}{Development of a voice virtual assistant for the
  geospatial data visualization application on the web}.
\newblock \bibinfo{journal}{ISPRS International Journal of Geo-Information}
  \bibinfo{volume}{12}, \bibinfo{pages}{441}.
\bibitem[{Mandal et~al.(2024)Mandal, Zou, Wilkho, Baig, Abedin, Zhou, Cai,
  Gharaibeh and Lam}]{mandal2024prime}
\bibinfo{author}{Mandal, D.}, \bibinfo{author}{Zou, L.},
  \bibinfo{author}{Wilkho, R.S.}, \bibinfo{author}{Baig, F.},
  \bibinfo{author}{Abedin, J.}, \bibinfo{author}{Zhou, B.},
  \bibinfo{author}{Cai, H.}, \bibinfo{author}{Gharaibeh, N.},
  \bibinfo{author}{Lam, N.}, \bibinfo{year}{2024}.
\newblock \bibinfo{title}{Prime: A cybergis platform for resilience inference
  measurement and enhancement}.
\newblock \bibinfo{journal}{Computers, Environment and Urban Systems}
  \bibinfo{volume}{114}, \bibinfo{pages}{102197}.
\bibitem[{Mansourian and Oucheikh(2024)}]{mansourian2024chatgeoai}
\bibinfo{author}{Mansourian, A.}, \bibinfo{author}{Oucheikh, R.},
  \bibinfo{year}{2024}.
\newblock \bibinfo{title}{Chatgeoai: Enabling geospatial analysis for public
  through natural language, with large language models}.
\newblock \bibinfo{journal}{ISPRS International Journal of Geo-Information}
  \bibinfo{volume}{13}, \bibinfo{pages}{348}.
\bibitem[{Manuardi(2024)}]{manuardi2024images}
\bibinfo{author}{Manuardi, D.}, \bibinfo{year}{2024}.
\newblock \bibinfo{title}{From Images to Code: Leveraging Computer Vision and
  Large Language Models for Front-End Automation}.
\newblock Ph.D. thesis. Politecnico di Torino.
\bibitem[{McBreen(2002)}]{mcbreen2002software}
\bibinfo{author}{McBreen, P.}, \bibinfo{year}{2002}.
\newblock \bibinfo{title}{Software craftsmanship: The new imperative}.
\newblock \bibinfo{publisher}{Addison-Wesley Professional}.
\bibitem[{Mendoza~Juan(2024)}]{mendoza2024development}
\bibinfo{author}{Mendoza~Juan, Y.}, \bibinfo{year}{2024}.
\newblock \bibinfo{title}{Development of a multi-agent, LLM-driven system to
  enhance human-machine interaction: integrating DSPy with modular agentic
  strategies and logical reasoning layers for the autonomous generation of
  smart contracts}.
\newblock Master's thesis. Universitat Polit{\`e}cnica de Catalunya.
\bibitem[{Meyer et~al.(2023)Meyer, Stadler, Frey, Radtke, Junghanns, Meissner,
  Dziwis, Bulert and Martin}]{meyer2023llm}
\bibinfo{author}{Meyer, L.P.}, \bibinfo{author}{Stadler, C.},
  \bibinfo{author}{Frey, J.}, \bibinfo{author}{Radtke, N.},
  \bibinfo{author}{Junghanns, K.}, \bibinfo{author}{Meissner, R.},
  \bibinfo{author}{Dziwis, G.}, \bibinfo{author}{Bulert, K.},
  \bibinfo{author}{Martin, M.}, \bibinfo{year}{2023}.
\newblock \bibinfo{title}{{Llm-assisted knowledge graph engineering:
  Experiments with chatgpt}}, in: \bibinfo{booktitle}{Working conference on
  Artificial Intelligence Development for a Resilient and Sustainable
  Tomorrow}, \bibinfo{organization}{Springer Fachmedien Wiesbaden Wiesbaden}.
  pp. \bibinfo{pages}{103--115}.
\bibitem[{Muste et~al.(2017)Muste, Carson, Xu and Mocanu}]{muste2017community}
\bibinfo{author}{Muste, M.}, \bibinfo{author}{Carson, A.}, \bibinfo{author}{Xu,
  H.}, \bibinfo{author}{Mocanu, M.}, \bibinfo{year}{2017}.
\newblock \bibinfo{title}{Community engagement in water resources planning
  using serious gaming}, in: \bibinfo{booktitle}{2017 13th IEEE International
  Conference on Intelligent Computer Communication and Processing (ICCP)},
  \bibinfo{organization}{IEEE}. pp. \bibinfo{pages}{445--451}.
\bibitem[{Nguyen-Duc et~al.(2023)Nguyen-Duc, Cabrero-Daniel, Przybylek, Arora,
  Khanna, Herda, Rafiq, Melegati, Guerra, Kemell et~al.}]{nguyen2023generative}
\bibinfo{author}{Nguyen-Duc, A.}, \bibinfo{author}{Cabrero-Daniel, B.},
  \bibinfo{author}{Przybylek, A.}, \bibinfo{author}{Arora, C.},
  \bibinfo{author}{Khanna, D.}, \bibinfo{author}{Herda, T.},
  \bibinfo{author}{Rafiq, U.}, \bibinfo{author}{Melegati, J.},
  \bibinfo{author}{Guerra, E.}, \bibinfo{author}{Kemell, K.K.}, et~al.,
  \bibinfo{year}{2023}.
\newblock \bibinfo{title}{Generative artificial intelligence for software
  engineering--a research agenda}.
\newblock \bibinfo{journal}{arXiv preprint arXiv:2310.18648} .
\bibitem[{Niloofar et~al.(2023)Niloofar, Lazarova-Molnar, Omitaomu, Xu and
  Li}]{niloofar2023general}
\bibinfo{author}{Niloofar, P.}, \bibinfo{author}{Lazarova-Molnar, S.},
  \bibinfo{author}{Omitaomu, F.}, \bibinfo{author}{Xu, H.},
  \bibinfo{author}{Li, X.}, \bibinfo{year}{2023}.
\newblock \bibinfo{title}{A general framework for human-in-the-loop cognitive
  digital twins}, in: \bibinfo{booktitle}{2023 Winter Simulation Conference
  (WSC)}, \bibinfo{organization}{IEEE}. pp. \bibinfo{pages}{3202--3213}.
\bibitem[{Parashar et~al.(2019)Parashar, Simonet, Rodero, Ghahramani, Agnew,
  Jantz and Honavar}]{parashar2019virtual}
\bibinfo{author}{Parashar, M.}, \bibinfo{author}{Simonet, A.},
  \bibinfo{author}{Rodero, I.}, \bibinfo{author}{Ghahramani, F.},
  \bibinfo{author}{Agnew, G.}, \bibinfo{author}{Jantz, R.},
  \bibinfo{author}{Honavar, V.}, \bibinfo{year}{2019}.
\newblock \bibinfo{title}{The virtual data collaboratory: A regional
  cyberinfrastructure for collaborative data-driven research}.
\newblock \bibinfo{journal}{Computing in Science \& Engineering}
  \bibinfo{volume}{22}, \bibinfo{pages}{79--92}.
\bibitem[{Phan et~al.(2024)Phan, Nguyen and Bui}]{phan2024hyperagent}
\bibinfo{author}{Phan, H.N.}, \bibinfo{author}{Nguyen, P.X.},
  \bibinfo{author}{Bui, N.D.}, \bibinfo{year}{2024}.
\newblock \bibinfo{title}{Hyperagent: Generalist software engineering agents to
  solve coding tasks at scale}.
\newblock \bibinfo{journal}{arXiv preprint arXiv:2409.16299} .
\bibitem[{Sandberg and Zhang(2024)}]{sandberg2024evaluating}
\bibinfo{author}{Sandberg, J.}, \bibinfo{author}{Zhang, Y.},
  \bibinfo{year}{2024}.
\newblock \bibinfo{title}{Evaluating the capabilities of gpt-4 in full-stack
  web development: A practical approach}.
\bibitem[{Shah and Al-Mohammad(2024)}]{shah2024optimizing}
\bibinfo{author}{Shah, H.}, \bibinfo{author}{Al-Mohammad, R.},
  \bibinfo{year}{2024}.
\newblock \bibinfo{title}{Optimizing software validation efficiency and
  scalability through mass parallel testing techniques in complex development
  environments}.
\newblock \bibinfo{journal}{International Journal of Intelligent Automation and
  Computing} \bibinfo{volume}{7}, \bibinfo{pages}{90--123}.
\bibitem[{Shanjun et~al.(2024)Shanjun, Pengpeng, Haoyuan, Jinchuan, Mei and
  Huazhou}]{shanjun2024design}
\bibinfo{author}{Shanjun, M.}, \bibinfo{author}{Pengpeng, Z.},
  \bibinfo{author}{Haoyuan, Z.}, \bibinfo{author}{Jinchuan, C.},
  \bibinfo{author}{Mei, L.}, \bibinfo{author}{Huazhou, C.},
  \bibinfo{year}{2024}.
\newblock \bibinfo{title}{Design and key technology research of industrial
  geographic information system}.
\newblock \bibinfo{journal}{National Remote Sensing Bulletin}
  \bibinfo{volume}{28}, \bibinfo{pages}{1189--1205}.
\bibitem[{Shao et~al.(2022)Shao, Wang, Berres, Yoshioka, Cook and
  Xu}]{shao2022computer}
\bibinfo{author}{Shao, Y.}, \bibinfo{author}{Wang, C.},
  \bibinfo{author}{Berres, A.}, \bibinfo{author}{Yoshioka, J.},
  \bibinfo{author}{Cook, A.}, \bibinfo{author}{Xu, H.}, \bibinfo{year}{2022}.
\newblock \bibinfo{title}{Computer vision-enabled smart traffic monitoring for
  sustainable transportation management}, in: \bibinfo{booktitle}{International
  Conference on Transportation and Development 2022}, pp.
  \bibinfo{pages}{34--45}.
\bibitem[{Siddiqui and Mead(2024)}]{siddiqui2024digital}
\bibinfo{author}{Siddiqui, F.M.}, \bibinfo{author}{Mead, C.J.},
  \bibinfo{year}{2024}.
\newblock \bibinfo{title}{Digital twin conops as a platform for airport master
  planning}, in: \bibinfo{booktitle}{2024 New Trends in Civil Aviation (NTCA)},
  \bibinfo{organization}{IEEE}. pp. \bibinfo{pages}{105--111}.
\bibitem[{Skarlatidou et~al.(2019)Skarlatidou, Hamilton, Vitos and
  Haklay}]{skarlatidou2019volunteers}
\bibinfo{author}{Skarlatidou, A.}, \bibinfo{author}{Hamilton, A.},
  \bibinfo{author}{Vitos, M.}, \bibinfo{author}{Haklay, M.},
  \bibinfo{year}{2019}.
\newblock \bibinfo{title}{What do volunteers want from citizen science
  technologies? a systematic literature review and best practice guidelines}.
\newblock \bibinfo{journal}{JCOM: Journal of Science Communication}
  \bibinfo{volume}{18}.
\bibitem[{Souffront~Alcantara et~al.(2018)Souffront~Alcantara, Kesler, Stealey,
  Nelson, Ames and Jones}]{souffront2018cyberinfrastructure}
\bibinfo{author}{Souffront~Alcantara, M.A.}, \bibinfo{author}{Kesler, C.},
  \bibinfo{author}{Stealey, M.J.}, \bibinfo{author}{Nelson, E.J.},
  \bibinfo{author}{Ames, D.P.}, \bibinfo{author}{Jones, N.L.},
  \bibinfo{year}{2018}.
\newblock \bibinfo{title}{Cyberinfrastructure and web apps for managing and
  disseminating the national water model}.
\newblock \bibinfo{journal}{JAWRA Journal of the American Water Resources
  Association} \bibinfo{volume}{54}, \bibinfo{pages}{859--871}.
\bibitem[{Steckler et~al.(2025)Steckler, Yu, Birdwell and Xu}]{steckler_2025}
\bibinfo{author}{Steckler, M.}, \bibinfo{author}{Yu, X.Y.},
  \bibinfo{author}{Birdwell, K.}, \bibinfo{author}{Xu, H.},
  \bibinfo{year}{2025}.
\newblock \bibinfo{title}{Five years of quality-controlled meteorological
  surface data at oak ridge reserve in tennessee}.
\newblock \URLprefix \url{https://doi.org/10.5281/zenodo.14744006},
  \DOIprefix\doi{10.5281/zenodo.14744006}.
\bibitem[{Sun et~al.(2023)Sun, Li, Huang, Yu and Long}]{sun2023gpt}
\bibinfo{author}{Sun, Y.X.}, \bibinfo{author}{Li, Z.M.},
  \bibinfo{author}{Huang, J.Z.}, \bibinfo{author}{Yu, N.z.},
  \bibinfo{author}{Long, X.}, \bibinfo{year}{2023}.
\newblock \bibinfo{title}{{GPT-4: the future of cosmetic procedure
  consultation?}}
\newblock \bibinfo{journal}{Aesthetic Surgery Journal} \bibinfo{volume}{43},
  \bibinfo{pages}{NP670--NP672}.
\bibitem[{Thakur et~al.(2020)Thakur, Sparks, Berres, Tansakul, Chinthavali,
  Whitehead, Schmidt, Xu, Fan, Spears et~al.}]{thakur2020covid}
\bibinfo{author}{Thakur, G.}, \bibinfo{author}{Sparks, K.},
  \bibinfo{author}{Berres, A.}, \bibinfo{author}{Tansakul, V.},
  \bibinfo{author}{Chinthavali, S.}, \bibinfo{author}{Whitehead, M.},
  \bibinfo{author}{Schmidt, E.}, \bibinfo{author}{Xu, H.},
  \bibinfo{author}{Fan, J.}, \bibinfo{author}{Spears, D.}, et~al.,
  \bibinfo{year}{2020}.
\newblock \bibinfo{title}{Covid-19 joint pandemic modeling and analysis
  platform}, in: \bibinfo{booktitle}{Proceedings of the 1st ACM SIGSPATIAL
  International Workshop on Modeling and Understanding the Spread of COVID-19},
  pp. \bibinfo{pages}{43--52}.
\bibitem[{Tupayachi et~al.(2024)Tupayachi, Xu, Omitaomu, Camur, Sharmin and
  Li}]{tupayachi2024towards}
\bibinfo{author}{Tupayachi, J.}, \bibinfo{author}{Xu, H.},
  \bibinfo{author}{Omitaomu, O.A.}, \bibinfo{author}{Camur, M.C.},
  \bibinfo{author}{Sharmin, A.}, \bibinfo{author}{Li, X.},
  \bibinfo{year}{2024}.
\newblock \bibinfo{title}{Towards next-generation urban decision support
  systems through ai-powered construction of scientific ontology using large
  language models—a case in optimizing intermodal freight transportation}.
\newblock \bibinfo{journal}{Smart Cities} \bibinfo{volume}{7},
  \bibinfo{pages}{2392--2421}.
\bibitem[{Xia et~al.(2024)Xia, Deng, Dunn and Zhang}]{xia2024agentless}
\bibinfo{author}{Xia, C.S.}, \bibinfo{author}{Deng, Y.}, \bibinfo{author}{Dunn,
  S.}, \bibinfo{author}{Zhang, L.}, \bibinfo{year}{2024}.
\newblock \bibinfo{title}{Agentless: Demystifying llm-based software
  engineering agents}.
\newblock \bibinfo{journal}{arXiv preprint arXiv:2407.01489} .
\bibitem[{Xiao et~al.(2024)Xiao, Chen, Li, Chen, Sun and
  Zhou}]{xiao2024prototype2code}
\bibinfo{author}{Xiao, S.}, \bibinfo{author}{Chen, Y.}, \bibinfo{author}{Li,
  J.}, \bibinfo{author}{Chen, L.}, \bibinfo{author}{Sun, L.},
  \bibinfo{author}{Zhou, T.}, \bibinfo{year}{2024}.
\newblock \bibinfo{title}{Prototype2code: End-to-end front-end code generation
  from ui design prototypes}, in: \bibinfo{booktitle}{International Design
  Engineering Technical Conferences and Computers and Information in
  Engineering Conference}, \bibinfo{organization}{American Society of
  Mechanical Engineers}. p. \bibinfo{pages}{V02BT02A038}.
\bibitem[{Xu et~al.(2022a)Xu, Berres, Liu, Allen-Dumas and
  Sanyal}]{xu2022overview}
\bibinfo{author}{Xu, H.}, \bibinfo{author}{Berres, A.}, \bibinfo{author}{Liu,
  Y.}, \bibinfo{author}{Allen-Dumas, M.R.}, \bibinfo{author}{Sanyal, J.},
  \bibinfo{year}{2022}a.
\newblock \bibinfo{title}{An overview of visualization and visual analytics
  applications in water resources management}.
\newblock \bibinfo{journal}{Environmental Modelling \& Software}
  \bibinfo{volume}{153}, \bibinfo{pages}{105396}.
\bibitem[{Xu et~al.(2023a)Xu, Berres, Shao, Wang, New and
  Omitaomu}]{xu2023toward}
\bibinfo{author}{Xu, H.}, \bibinfo{author}{Berres, A.}, \bibinfo{author}{Shao,
  Y.}, \bibinfo{author}{Wang, C.R.}, \bibinfo{author}{New, J.R.},
  \bibinfo{author}{Omitaomu, O.A.}, \bibinfo{year}{2023}a.
\newblock \bibinfo{title}{Toward a smart metaverse city: Immersive realism and
  3d visualization of digital twin cities}.
\newblock \bibinfo{journal}{Advances in Scalable and Intelligent Geospatial
  Analytics} , \bibinfo{pages}{245--257}.
\bibitem[{Xu et~al.(2021a)Xu, Berres, Tennille, Ravulaparthy, Wang and
  Sanyal}]{xu2021continuous}
\bibinfo{author}{Xu, H.}, \bibinfo{author}{Berres, A.},
  \bibinfo{author}{Tennille, S.A.}, \bibinfo{author}{Ravulaparthy, S.K.},
  \bibinfo{author}{Wang, C.}, \bibinfo{author}{Sanyal, J.},
  \bibinfo{year}{2021}a.
\newblock \bibinfo{title}{Continuous emulation and multiscale visualization of
  traffic flow using stationary roadside sensor data}.
\newblock \bibinfo{journal}{IEEE Transactions on Intelligent Transportation
  Systems} \bibinfo{volume}{23}, \bibinfo{pages}{10530--10541}.
\bibitem[{Xu et~al.(2021b)Xu, Berres, Thakur, Sanyal and
  Chinthavali}]{xu2021episemblevis}
\bibinfo{author}{Xu, H.}, \bibinfo{author}{Berres, A.},
  \bibinfo{author}{Thakur, G.}, \bibinfo{author}{Sanyal, J.},
  \bibinfo{author}{Chinthavali, S.}, \bibinfo{year}{2021}b.
\newblock \bibinfo{title}{Episemblevis: A geo-visual analysis and comparison of
  the prediction ensembles of multiple covid-19 models}.
\newblock \bibinfo{journal}{Journal of biomedical informatics}
  \bibinfo{volume}{124}, \bibinfo{pages}{103941}.
\bibitem[{Xu et~al.(2021c)Xu, Berres, Wang, LaClair and
  Sanyal}]{xu2021visualizing}
\bibinfo{author}{Xu, H.}, \bibinfo{author}{Berres, A.}, \bibinfo{author}{Wang,
  C.R.}, \bibinfo{author}{LaClair, T.J.}, \bibinfo{author}{Sanyal, J.},
  \bibinfo{year}{2021}c.
\newblock \bibinfo{title}{Visualizing vehicle acceleration and braking energy
  at intersections along a major traffic corridor}, in:
  \bibinfo{booktitle}{Proceedings of the Twelfth ACM International Conference
  on Future Energy Systems}, pp. \bibinfo{pages}{401--405}.
\bibitem[{Xu et~al.(2023b)Xu, Berres, Yoginath, Sorensen, Nugent, Severino,
  Tennille, Moore, Jones and Sanyal}]{xu2023smart}
\bibinfo{author}{Xu, H.}, \bibinfo{author}{Berres, A.},
  \bibinfo{author}{Yoginath, S.B.}, \bibinfo{author}{Sorensen, H.},
  \bibinfo{author}{Nugent, P.J.}, \bibinfo{author}{Severino, J.},
  \bibinfo{author}{Tennille, S.A.}, \bibinfo{author}{Moore, A.},
  \bibinfo{author}{Jones, W.}, \bibinfo{author}{Sanyal, J.},
  \bibinfo{year}{2023}b.
\newblock \bibinfo{title}{Smart mobility in the cloud: Enabling real-time
  situational awareness and cyber-physical control through a digital twin for
  traffic}.
\newblock \bibinfo{journal}{IEEE Transactions on Intelligent Transportation
  Systems} \bibinfo{volume}{24}, \bibinfo{pages}{3145--3156}.
\bibitem[{Xu et~al.(2024a)Xu, Boyaci, Lian and Wilson}]{xu2024explainable}
\bibinfo{author}{Xu, H.}, \bibinfo{author}{Boyaci, A.}, \bibinfo{author}{Lian,
  J.}, \bibinfo{author}{Wilson, A.}, \bibinfo{year}{2024}a.
\newblock \bibinfo{title}{Explainable ai for multivariate time series pattern
  exploration: Latent space visual analytics with temporal fusion transformer
  and variational autoencoders in power grid event diagnosis}.
\newblock \bibinfo{journal}{arXiv e-prints} , \bibinfo{pages}{arXiv--2412}.
\bibitem[{Xu et~al.(2019a)Xu, Demir, Koylu and Muste}]{xu2019web}
\bibinfo{author}{Xu, H.}, \bibinfo{author}{Demir, I.}, \bibinfo{author}{Koylu,
  C.}, \bibinfo{author}{Muste, M.}, \bibinfo{year}{2019}a.
\newblock \bibinfo{title}{A web-based geovisual analytics platform for
  identifying potential contributors to culvert sedimentation}.
\newblock \bibinfo{journal}{Science of the Total Environment}
  \bibinfo{volume}{692}, \bibinfo{pages}{806--817}.
\bibitem[{Xu et~al.(2024b)Xu, Li, Tupayachi, Lian and
  Omitaomu}]{xu2024automating}
\bibinfo{author}{Xu, H.}, \bibinfo{author}{Li, X.}, \bibinfo{author}{Tupayachi,
  J.}, \bibinfo{author}{Lian, J.J.}, \bibinfo{author}{Omitaomu, O.A.},
  \bibinfo{year}{2024}b.
\newblock \bibinfo{title}{Automating bibliometric analysis with sentence
  transformers and retrieval-augmented generation (rag): A pilot study in
  semantic and contextual search for customized literature characterization for
  high-impact urban research}, in: \bibinfo{booktitle}{Proceedings of the 2nd
  ACM SIGSPATIAL International Workshop on Advances in Urban-AI}, pp.
  \bibinfo{pages}{43--49}.
\bibitem[{Xu et~al.(2022b)Xu, Liu, Lian, Li, Malhotra and Omitaomu}]{xu2022geo}
\bibinfo{author}{Xu, H.}, \bibinfo{author}{Liu, X.}, \bibinfo{author}{Lian,
  J.}, \bibinfo{author}{Li, Y.}, \bibinfo{author}{Malhotra, M.},
  \bibinfo{author}{Omitaomu, O.A.}, \bibinfo{year}{2022}b.
\newblock \bibinfo{title}{A geo-visual analysis for exploring the socioeconomic
  benefits of the heating electrification using geothermal energy}, in:
  \bibinfo{booktitle}{Proceedings of the 5th ACM SIGSPATIAL International
  Workshop on Advances in Resilient and Intelligent Cities}, pp.
  \bibinfo{pages}{11--15}.
\bibitem[{Xu et~al.(2019b)Xu, Muste and Demir}]{xu2019web2}
\bibinfo{author}{Xu, H.}, \bibinfo{author}{Muste, M.}, \bibinfo{author}{Demir,
  I.}, \bibinfo{year}{2019}b.
\newblock \bibinfo{title}{Web-based geospatial platform for the analysis and
  forecasting of sedimentation at culverts}.
\newblock \bibinfo{journal}{Journal of Hydroinformatics} \bibinfo{volume}{21},
  \bibinfo{pages}{1064--1081}.
\bibitem[{Xu et~al.(2024c)Xu, Shao, Chen, Wang and Berres}]{xu2024semi}
\bibinfo{author}{Xu, H.}, \bibinfo{author}{Shao, Y.}, \bibinfo{author}{Chen,
  J.}, \bibinfo{author}{Wang, C.}, \bibinfo{author}{Berres, A.},
  \bibinfo{year}{2024}c.
\newblock \bibinfo{title}{Semi-automatic geographic information system
  framework for creating photo-realistic digital twin cities to support
  autonomous driving research}.
\newblock \bibinfo{journal}{Transportation research record} ,
  \bibinfo{pages}{03611981231205884}.
\bibitem[{Xu et~al.(2022c)Xu, Wang, Berres, LaClair and
  Sanyal}]{xu2022interactive}
\bibinfo{author}{Xu, H.}, \bibinfo{author}{Wang, C.}, \bibinfo{author}{Berres,
  A.}, \bibinfo{author}{LaClair, T.}, \bibinfo{author}{Sanyal, J.},
  \bibinfo{year}{2022}c.
\newblock \bibinfo{title}{Interactive web application for traffic simulation
  data management and visualization}.
\newblock \bibinfo{journal}{Transportation research record}
  \bibinfo{volume}{2676}, \bibinfo{pages}{274--292}.
\bibitem[{Xu et~al.(2020)Xu, Windsor, Muste and Demir}]{xu2020web}
\bibinfo{author}{Xu, H.}, \bibinfo{author}{Windsor, M.},
  \bibinfo{author}{Muste, M.}, \bibinfo{author}{Demir, I.},
  \bibinfo{year}{2020}.
\newblock \bibinfo{title}{A web-based decision support system for collaborative
  mitigation of multiple water-related hazards using serious gaming}.
\newblock \bibinfo{journal}{Journal of environmental management}
  \bibinfo{volume}{255}, \bibinfo{pages}{109887}.
\bibitem[{Xu et~al.(2023c)Xu, Yuan, Berres, Shao, Wang, Li, LaClair, Sanyal and
  Wang}]{xu2023mobile}
\bibinfo{author}{Xu, H.}, \bibinfo{author}{Yuan, J.}, \bibinfo{author}{Berres,
  A.}, \bibinfo{author}{Shao, Y.}, \bibinfo{author}{Wang, C.},
  \bibinfo{author}{Li, W.}, \bibinfo{author}{LaClair, T.J.},
  \bibinfo{author}{Sanyal, J.}, \bibinfo{author}{Wang, H.},
  \bibinfo{year}{2023}c.
\newblock \bibinfo{title}{A mobile edge computing framework for traffic
  optimization at urban intersections through cyber-physical integration}.
\newblock \bibinfo{journal}{IEEE Transactions on Intelligent Vehicles} .
\bibitem[{Yu et~al.(2021)Yu, Ibarra, Kumar and Chergarova}]{yu2021coevolution}
\bibinfo{author}{Yu, Y.}, \bibinfo{author}{Ibarra, J.E.},
  \bibinfo{author}{Kumar, K.}, \bibinfo{author}{Chergarova, V.},
  \bibinfo{year}{2021}.
\newblock \bibinfo{title}{Coevolution of cyberinfrastructure development and
  scientific progress}.
\newblock \bibinfo{journal}{Technovation} \bibinfo{volume}{100},
  \bibinfo{pages}{102180}.
\bibitem[{Zhang et~al.(2023)Zhang, Chen, Shen, Ding, Tenenbaum and
  Gan}]{zhang2023planning}
\bibinfo{author}{Zhang, S.}, \bibinfo{author}{Chen, Z.}, \bibinfo{author}{Shen,
  Y.}, \bibinfo{author}{Ding, M.}, \bibinfo{author}{Tenenbaum, J.B.},
  \bibinfo{author}{Gan, C.}, \bibinfo{year}{2023}.
\newblock \bibinfo{title}{Planning with large language models for code
  generation}.
\newblock \bibinfo{journal}{arXiv preprint arXiv:2303.05510} .
\bibitem[{Zhang et~al.(2024)Zhang, Wang, He, Li, Mai, Lin, Wei and
  Yu}]{zhang2024bb}
\bibinfo{author}{Zhang, Y.}, \bibinfo{author}{Wang, Z.}, \bibinfo{author}{He,
  Z.}, \bibinfo{author}{Li, J.}, \bibinfo{author}{Mai, G.},
  \bibinfo{author}{Lin, J.}, \bibinfo{author}{Wei, C.}, \bibinfo{author}{Yu,
  W.}, \bibinfo{year}{2024}.
\newblock \bibinfo{title}{Bb-geogpt: A framework for learning a large language
  model for geographic information science}.
\newblock \bibinfo{journal}{Information Processing \& Management}
  \bibinfo{volume}{61}, \bibinfo{pages}{103808}.

\end{thebibliography}

\end{document}